\documentclass{article}

\PassOptionsToPackage{numbers, compress}{natbib}
\usepackage[preprint]{neurips_2026}

\usepackage[utf8]{inputenc}
\usepackage[T1]{fontenc}
\usepackage{hyperref}
\usepackage{url}
\usepackage{booktabs}
\usepackage{amsfonts}
\usepackage{amsmath,amssymb}
\usepackage{nicefrac}
\usepackage{microtype}
\usepackage{xcolor}
\usepackage{algorithm}
\usepackage{algorithmic}
\usepackage{enumitem}
\usepackage{graphicx}
\usepackage{multirow}
\usepackage{tabularx}
\usepackage{array}
\usepackage{float}
\usepackage{mathtools}

\usepackage{tikz}
\usetikzlibrary{arrows.meta, positioning, calc, shapes.geometric, fit, backgrounds}

\newcolumntype{L}[1]{>{\raggedright\arraybackslash}p{#1}}
\newcolumntype{C}[1]{>{\centering\arraybackslash}p{#1}}

\DeclareMathOperator*{\argmax}{argmax}

\newcommand{\theoryopt}[1]{}

\title{The Context Gathering Decision Process: \\A POMDP Framework for Agentic Search}

\author{%
 Chinmaya Kausik\thanks{Work initiated as an intern at Netflix.} \\
  University of Michigan\\
  \texttt{ckausik@umich.edu} \\
  \And 
  Adith Swaminathan \\
  Netflix\\
  \texttt{aswaminathan@netflix.com} \\
  \And 
  Nathan Kallus \\
  Netflix\\
  \texttt{nkallus@netflix.com}
}

\begin{document}

\maketitle

\begin{abstract}

Large Language Model (LLM) agents are deployed in complex environments -- such as massive codebases, enterprise databases, and conversational histories -- where the relevant state far exceeds their context windows. 
To navigate these spaces, an agent must iteratively explore the environment to find relevant information. 
However, without explicit infrastructure, an agent's working memory can degrade into lossy representations of the search state, resulting in redundant work (e.g. repetitive looping) and premature stopping. 
In this work, we formalize this challenge as the Context Gathering Decision Process (CGDP), a specialized Partially Observable Markov Decision Process, where an agent's objective is to adaptively refine its belief state to isolate the necessary information for a task. 
We model an LLM's behavior as approximate Thompson Sampling within this CGDP, and introduce a predicate-based method that decomposes an LLM's implicit search into explicit and modular operations. 
We then derive two plug-and-play interventions for iterative LLM agents: a persistent, predicate-based belief state that bounds context while preserving multi-hop reasoning, and a programmatic exhaustion gate that halts unproductive search without premature stopping. 
Across four methods and three question-answering domains, we empirically validate that replacing an LLM's implicit state with our CGDP-motivated belief state improves multi-hop reasoning by up to $11.4\%$; while the modular programmatic exhaustion detection saves up to $39\%$ of tokens without any degradation in agent performance. 
Ultimately, we argue that framing the LLM agent loop as a CGDP can guide the design of modular, non-interfering improvements to agentic search harnesses. 

\end{abstract}

\section{Introduction}\label{sec:intro}

Large Language Model (LLM) agents are increasingly deployed on complex, real-world environments where the relevant state far exceeds their reliable working context. Deep-research agents search the web~\citep{zheng-etal-2025-deepresearcher}; coding assistants search repositories and run shell tools~\citep{swe-agent-2024}; support agents retrieve from enterprise knowledge bases~\citep{lewis2020retrievalaugmented}; all these systems face the same fundamental challenge: the agent cannot load the entire environment into its prompt. Instead, it must iteratively interact with an observation function -- such as a Python REPL, a search engine API, or a vector database -- to gather the necessary information. 

Despite active research into extending LLM context windows~\citep{li-etal-2024-loogle, wu-etal-2024-extending},  
long-context models still exhibit position sensitivity~\citep{liu2023lost}, degrade with hard negatives~\citep{jin2025longcontextrag}, are unreliable in multi-turn interaction~\citep{laban2026llmsgetlost}, and stop prematurely during long-horizon search~\citep{yen2025lostmaze}. Thus, building agentic harnesses -- infrastructure that allows the LLM to actively manage and search external context -- has been studied extensively~\citep{shao2023enhancing, trivedi2023ircot,yao2023react,packer2024memgptllmsoperatingsystems,rozanov2024stateact}. 

Agentic harnesses typically append raw observations directly to the LLM prompt~\citep{shao2023enhancing,trivedi2023ircot,yao2023react}; over long horizons, this implicit state tracking leads to severe failures: agents lose track of their original objective, fall into repetitive query loops, hallucinate parametric knowledge instead of corpus evidence, and fail to reliably recognize when a search is completely exhausted. For instance, a coding agent that is resolving a software bug may repeatedly retrieve the same file, oscillating between identical hypotheses without realizing that its search has stagnated. 

To address these failures, we formalize the interactive information-seeking agent loop as the Context Gathering Decision Process (CGDP). The CGDP is a specialized Partially Observable Markov Decision Process (POMDP)~\citep{POMDP_1998} where the hidden state is the vast external corpus, actions are tool calls, observations are retrieved information, 
and the objective is to adaptively refine the agent's belief state to identify task-relevant information from the hidden state. As shown in Table~\ref{tab:cgdp_applications}, this mathematical framework abstracts the details of many modern agents.

\begin{table}[htb]
    \centering
    \caption{Across diverse agentic applications, the underlying challenge remains to navigate a massive hidden state via a constrained observation function to satisfy a user query, which we model as a Context Gathering Decision Process (CGDP).}
    \label{tab:cgdp_applications}
    \begin{tabular}{@{}lll@{}}
        \toprule
        \textbf{Application} & \textbf{Hidden State} & \textbf{Action Space (Observation Function)} \\
        \midrule
        \textbf{Code Assistants}    & Codebase                          & Linux File System \& BASH Commands            \\
        \textbf{Agentic Memory}     & Conversational History                  & Vector Database Retrieval                    \\
        \textbf{Support Agents / RAG}& Knowledge-Base                & SQL / BM25 / Embedding Queries            \\
        \textbf{Recursive LMs}      & Python Data Objects             & Python REPL \& Code Interpreter                   \\
        \textbf{Deep Research}      & World Wide Web              & Search Engine APIs (e.g. Google)      \\
        \bottomrule
    \end{tabular}
\end{table}

Through the lens of CGDP, we model an LLM's behavior as approximate Thompson Sampling~\citep{thompson1933,russo2018tutorial,osband_thompson_sampling}, where the model implicitly samples a hypothesis and takes information-gathering actions conditioned on it. To make this process explicit, we introduce Predicate-Based Adaptive Identification (PBAI), an abstract algorithm that decomposes agentic search into modular operations: assess stopping, select action, observe, and update belief. 
By mapping state-of-the-art agent harnesses to PBAI, we can pinpoint precisely where their implicit reasoning lacks the necessary infrastructure for reliable, long-horizon search. 

Based on our framework, we derive two modular interventions for agentic search harnesses:
\begin{enumerate}[leftmargin=*,nosep]
\item The Predicate-Based Belief State: A persistent, explicitly managed data structure that forces the agent to extract findings and track open questions iteratively, bounding the context footprint while preserving multi-hop reasoning performance.
\item The Programmatic Exhaustion Gate: A stopping mechanism that halts unproductive search. Rather than relying on an LLM's self-assessment (which may be poorly calibrated~\citep{guo2024calibration,huang2024selfcorrect}), the mechanism uses programmatic signals like action similarity and observation novelty to prevent redundant looping and premature stopping.
\end{enumerate}

We empirically validate these interventions across four agent search harnesses and three domains (multi-session conversational Question-Answering, multi-hop Wikipedia QA, and code-repository QA). We find that replacing an LLM's implicit search state with the predicate-based belief state never degrades agent performance and improves multi-hop reasoning by up to $11.4\%$. Furthermore, the programmatic exhaustion gate safely reduces token consumption by up to $39\%$ on stateful harnesses without sacrificing task accuracy. 
Ultimately, we demonstrate that formalizing LLM agentic search as a CGDP provides a blueprint for designing reliable modular harness improvements. 

\section{Related Work}\label{sec:related}

Our contributions are at the intersection of retrieval-augmented generation, long-horizon agentic search, and LLM meta-cognition. 
While prior work has empirically improved specific components of the agent loop, we provide a unifying framework to understand how these components interact. 

\paragraph{Iterative RAG and Agent Harnesses.} Multi-round methods generally accumulate state implicitly through a growing context window. IRCoT~\citep{trivedi2023ircot} interleaves retrieval with chain-of-thought reasoning, Iter-RetGen~\citep{shao2023enhancing} uses previous generations to guide subsequent retrieval queries, and ReAct~\citep{yao2023react} tracks history through structured Thought-Action-Observation trajectories. Although these methods perform agentic search, their implicit state tracking frequently degrades over long episodes. More recent approaches introduce explicit per-retrieval interventions: Self-RAG~\citep{asai2024selfrag} adds retrieval-time self-reflection, Corrective RAG~\citep{yan2024correctiverag} corrects retrieval queries based on relevance assessment and FAIR-RAG~\citep{aghajani2025fairrag} introduces gap analysis via an evidence checklist. 
Similarly, MemGPT~\citep{packer2024memgptllmsoperatingsystems} provides LLMs with explicit memory management tools. 
While approaches like StateAct~\citep{rozanov2024stateact} or FAIR-RAG~\citep{aghajani2025fairrag} rely on LLM-managed summaries or checklists, our framework demonstrates that \emph{orchestrator-enforced}, strictly curated belief states outperform an LLM's self-managed tool use or memory edits. 

\paragraph{Long-Horizon Search and Corpus Organization.} To address unbounded context, offline memory organization methods like A-MEM~\citep{yu2025amem}, HippoRAG~\citep{hipporag2024}, HopRAG~\citep{hoprag2025} and GraphRAG~\citep{graphrag2024} structure the underlying corpus into graphs to facilitate easier retrieval, whereas in our work we organize the agent's understanding of what it has found from the corpus online. For long-horizon online search, SLIM~\citep{yen2025lostmaze} separates search from browsing and periodically summarizes trajectories to manage context, while 
AggAgent~\citep{lee2026agenticaggregation} generates parallel retrieval trajectories and synthesizes them on demand. 
These systems support our core premise: reliability improves when the external context is actively managed rather than passively appended. 

\paragraph{Stopping Criteria and Metacognition.} A critical challenge in iterative search is knowing when to stop. To detect whether to initiate retrieval, there are models such as FLARE~\citep{jiang2023flare} (using token-level confidence scores) and DRAGIN~\citep{su2024dragin} (using attention-based scores). 
In contrast, our work addresses post-retrieval stagnation. 
Relying on LLMs to self-assess stopping criteria has proven brittle; recent evidence shows that LLMs cannot reliably self-correct reasoning without external feedback~\citep{huang2024selfcorrect} and that their verbalized confidence is poorly calibrated~\citep{guo2024calibration}. 
Motivated by this, our programmatic exhaustion gate replaces LLM self-assessment with heuristic stagnation signals and improves token efficiency without degrading search accuracy.

\section{Framework}\label{sec:framework}

To understand why iterative LLM agents often fail at long-horizon search, we seek to separate the formulation of information-seeking problems from the specific prompt engineering used to solve them. In this section, we formalize agentic search as a specific decision-making process, define the notion of success, and diagnose why LLMs become sub-optimal agents for this process.

\subsection{The Context Gathering Decision Process}\label{sec:cgdp}

A Context Gathering Decision Process (CGDP) can be viewed as a POMDP~\citep{POMDP_1998} with terminal rewards and per-action costs, closely related to sequential identification~\citep{pmlr-v49-russo16, pmlr-v49-garivier16a}; it is defined by a task $q$ (e.g. user query), a massive hidden world state $c \in \mathcal{C}$ (e.g. codebase), an action space $\mathcal{A}$ (e.g. LLM-callable tools), an observation function $F: \mathcal{A} \times \mathcal{C} \to \mathcal{O}$ that maps an agent's actions to observable text chunks, 
and a per-action cost $\lambda$. 
At each timestep $t$, the agent selects an action $a_t \in \mathcal{A}$ which can either be an environment query (e.g. a BASH command) that incurs cost $Cost(a_t)$ and creates observation $o_t = F(a_t, c)$, or a termination action that returns a final answer $a_{\text{final}}$. 
The environment evaluates the agent's final answer via a binary success function $Success(q, c, a_{\text{final}}) \in \{ 0, 1\}$. 
Let $a^*(q,c)$ denote an optimal answer, i.e., $Success(q, c, a^*(q,c))=1$. 

Prompted with query $q$ alone, an LLM lacks the context to produce $a^*(q,c)$ and will abstain or hallucinate. 
Unable to process the entire hidden state $c$ at once, it must iteratively interact with $F$ to gather a sufficient subset of information. 

The optimal CGDP agent maximizes expected success while minimizing exploration cost (e.g. LLM token budget and/or latency) across a task distribution $\mathcal{D}$:
\begin{equation}
\argmax_{\mathrm{Policy}} \mathbb{E}_{(q,c) \sim \mathcal{D}} \left[ \mathrm{Success}(q, c, \mathrm{Policy}(q, c)) - \lambda \sum_{t=1}^T \mathrm{Cost}(a_t) \right].
\end{equation}

Crucially, the true state of the environment includes $c$, but the observation $o_t$ at each step is only the fragment of $c$ that the agent explicitly chose to observe via $F(a_t, c)$. Therefore, a successful CGDP agent must maintain an internal belief state $b_t$~\citep{POMDP_1998} that synthesizes its historical observations, tracks its progress towards $a^*(q,c)$ and guides the selection of the next action $a_t$. 

\subsection{LLMs' Behavior in CGDPs}\label{sec:llm_cgdp}

Agentic harnesses (e.g. ReAct~\citep{yao2023react}, IRCoT~\citep{trivedi2023ircot}) solve a CGDP by deploying an LLM as a policy, maintaining belief state implicitly by concatenating observation history. At step $t$, the state is 
\begin{equation}
\label{eq:naive_state}
    s_{t+1} = Truncate(s_t \oplus \{ a_t, o_t \} ),
\end{equation} 
where the truncation programmatically drops the older steps to obey the limits of LLM context windows. The policy is implemented simply through autoregressive generation $a_{t+1} = LLM(s_{t+1})$. Viewing LLM agents as CGDP policies reveals that they can suffer from failure modes because they lack fundamental mechanisms known to be beneficial for navigating POMDPs:
\begin{enumerate}[leftmargin=*,nosep]
\item Lossy Representation (Goal Forgetting): A vanilla LLM agent relies on the growing history $s_t$ as its belief state. As $t$ increases, the LLM must implicitly infer ``what is the goal?'' and ``what has been figured out so far?'' at every step. Empirical studies show that LLMs can 
ignore evidence in the middle of long contexts~\citep{liu2023lost}, which can cause goal drift.
\item Premature Stopping: Optimal agents in POMDPs stop exploring when the expected information gain of the next action is outweighed by its cost~\citep{pmlr-v49-russo16, pmlr-v49-garivier16a}. However, we conjecture that LLMs are trained on instruction-following datasets where the most rewarded behavior is to immediately produce an answer when a plausible one is found. Therefore, when the corpus $c$ contains irrelevant distractors or adversarial honeypots, the LLM agent can trigger premature termination~\citep{yen2025lostmaze}.
\item Insufficient Exploration: Many sequential decision-making algorithms explore via optimism~\citep{auer2002finite}, but LLMs do not have such mechanisms. When an LLM samples an uninformative observation $o_t$, it can repeatedly generate similar actions $a_{t+1} \approx a_t$ (mode collapse), without any structural awareness that its search has stagnated. 
\end{enumerate}

To overcome these LLM-specific failure modes, we next describe an abstracted algorithm for CGDPs that uses explicit state tracking and study how state-of-the-art LLM harnesses map to it.

\section{Abstract Algorithm}\label{sec:algorithm}

\begin{figure}[htbp]
  \centering
  \begin{minipage}[t]{0.58\textwidth}
    \vspace{0pt} 
    \resizebox{\linewidth}{!}{%
      \begin{tikzpicture}[
          x=1mm, y=1mm,
          font=\sffamily,
          every node/.style={inner sep=0pt, outer sep=0pt},
          arr/.style       = {-{Stealth[length=2mm,width=1.6mm]}, line width=0.45pt, draw=gray!75!black},
          arrblue/.style   = {-{Stealth[length=2mm,width=1.6mm]}, line width=0.55pt, draw=blue!70!black},
          arrorange/.style = {-{Stealth[length=2mm,width=1.6mm]}, line width=0.55pt, draw=orange!75!red},
          arrspine/.style  = {-{Stealth[length=1.7mm,width=1.4mm]}, line width=0.5pt, draw=gray!60},
          dashedblue/.style   = {arrblue, dash pattern=on 1.2mm off 0.8mm},
          dashedorange/.style = {arrorange, dash pattern=on 1mm off 0.6mm},
          dashedspine/.style  = {arrspine, dash pattern=on 1mm off 0.6mm},
          gatebox/.style       = {draw=orange!75!red, fill=orange!8, line width=0.5pt, rounded corners=1mm, minimum width=45mm, minimum height=7mm},
          beliefbox/.style     = {draw=blue!70!black, fill=blue!10, line width=0.5pt, rounded corners=1mm, minimum width=22mm, minimum height=12mm},
          methodbox/.style     = {draw=gray!60!black, fill=gray!4, line width=0.5pt, rounded corners=1mm, minimum width=16mm, minimum height=18mm},
          innerbox/.style      = {draw=gray!50, fill=white, line width=0.3pt, rounded corners=0.5mm, minimum width=12mm, minimum height=5mm},
          corpusbox/.style     = {draw=green!50!black, fill=green!12, line width=0.4pt, rounded corners=0.5mm, minimum width=12mm, minimum height=5mm},
          extractorbox/.style  = {draw=blue!70!black, fill=blue!10, line width=0.5pt, rounded corners=1mm, minimum width=45mm, minimum height=8mm},
          orchbox/.style       = {draw=blue!60, line width=0.35pt, rounded corners=1mm, dash pattern=on 1.2mm off 0.8mm, opacity=0.55},
          fixtag/.style        = {fill=#1, text=white, font=\bfseries\sffamily\tiny, rounded corners=0.3mm, inner xsep=1mm, inner ysep=0.4mm},
          stepname/.style      = {font=\bfseries\sffamily\scriptsize, text=black, anchor=west}
        ]

        \useasboundingbox (-8,0) rectangle (75,55);

        \node[font=\bfseries\sffamily\tiny, text=gray!40!black, anchor=west] at (1,53)  {PBAI};
        
        \node[stepname] (s1n) at (0, 48) {1. Stop?};
        \node[stepname] (s2n) at (0, 36) {2. Act};
        \node[stepname] (s3n) at (0, 24) {3. Observe};
        \node[stepname] (s4n) at (0, 12) {4. Update};

        \draw[arrspine] (2, 46) -- (2, 38);
        \draw[arrspine] (2, 34) -- (2, 26);
        \draw[arrspine] (2, 22) -- (2, 14);
        
        \draw[dashedspine] (-1, 10) -- (-6, 10) -- (-6, 50) -- (-1, 50);
        \node[font=\itshape\sffamily\tiny, text=gray!50!black, rotate=90, anchor=south] at (-6, 30) {next round};

        \draw[orchbox] (14, 5) rectangle (74, 51);
        \node[font=\bfseries\sffamily\tiny, text=blue!70!black, opacity=0.8, anchor=north east] at (73, 53) {ORCHESTRATOR};

        \node[gatebox] (gate) at (44, 46) {};
        \node[fixtag=orange!75!red] at (25, 48.5) {Gate};
        \node[font=\bfseries\sffamily\scriptsize, text=orange!50!black] at (44, 46) {Stagnation?};

        \draw[arrorange] (25, 50) -- (25, 54);
        \node[font=\itshape\sffamily\tiny, text=orange!50!black, anchor=west] at (26, 53) {stop $\to$ $a_{\text{final}}$ using $b_t$};

        \node[beliefbox] (sbox) at (28, 30) {};
        \node[fixtag=blue!70!black] at (25, 35) {Belief State};
        \node[font=\bfseries\itshape\sffamily\large, text=blue!50!black] at (28, 30.5) {$b_t$};
        \node[font=\sffamily\tiny, text=blue!60!black] at (28, 26) {Facts \& Questions};

        \node[methodbox] (mbox) at (49, 30) {};
        \node[font=\bfseries\sffamily\tiny, text=gray!50!black, anchor=north west] at (42, 38.5) {AGENT};
        
        \node[innerbox] (llm) at (50, 34) {\scriptsize LLM};
        \node[innerbox] (ret) at (50, 26) {\scriptsize Tools $F$};
        \draw[arr] (50, 31.5) -- (50, 28.5);
        \node[font=\itshape\sffamily\tiny, anchor=west] at (51, 30) {$a_t$};

        \node[corpusbox] (corp) at (65, 26) {\scriptsize Corpus $c$};
        \draw[arr, draw=green!50!black] (56, 26.5) -- (59, 26.5);
        \draw[arr, draw=green!50!black] (59, 25.5) -- (56, 25.5);

        \draw[arrblue] (39, 34) -- (43, 34);
        \node[font=\itshape\sffamily\tiny, text=blue!70!black] at (41, 35.5) {inject};

        \draw[arrblue] (54, 21) -- (54, 15);
        \node[font=\itshape\sffamily\tiny, text=blue!70!black, anchor=west] at (55, 18) {$o_t$};

        \node[extractorbox] (ext) at (44, 11) {};
        \node[fixtag=blue!70!black] at (25, 14) {Extractor};
        \node[font=\bfseries\sffamily\scriptsize, text=blue!50!black] at (44, 11) {Resolve $o_t \to b_{t+1}$};

        \draw[dashedblue] (28, 24) -- (28, 15);
        
        \draw[arrblue, line width=0.6pt] (21, 11) -- (15.5, 11) -- (15.5, 30) -- (17, 30);
        \node[font=\bfseries\itshape\sffamily\tiny, text=blue!70!black, rotate=90, anchor=south] at (15.5, 20.5) {write $b_{t+1}$};

        \draw[dashedorange] (72, 34) -- (72, 46) -- (66.5, 46);
        \draw[orange!75!red, line width=0.4pt, dash pattern=on 1mm off 0.6mm] (66.5, 11) -- (72, 11) -- (72, 34);
        \node[font=\itshape\sffamily\tiny, text=orange!55!black, rotate=-90, anchor=south] at (72, 26) {signals: $a_t, o_t$};

      \end{tikzpicture}%
    }
    \caption{The PBAI loop with our two harness interventions. The \textbf{Belief State} (blue) is updated by the \textbf{Extractor} and injected into the agent prompt. The \textbf{Gate} (orange) evaluates programmatic signals to detect stagnation.}
    \label{fig:architecture}
  \end{minipage}\hfill
  \begin{minipage}[t]{0.4\textwidth}
    \vspace{-14pt} 
    \begin{algorithm}[H]
      \caption{The PBAI Loop. Agent selects actions based on unresolved predicates, and updates belief state until the query is satisfied or the budget is exhausted.
      }
      \label{alg:ideal}
      \begin{algorithmic}[1]
        \REQUIRE Query $q$, budget $B$
        \STATE Initialize belief state $b_0$ from $q$
        \WHILE{$\text{cost} < B$}
          \STATE \textbf{1. Stop?:} If facts in $b_t$ satisfy $q \to$ \textbf{break the while loop}
          \STATE \textbf{2. Act:} Generate action $a_t$ targeting the top open predicate in $b_t$.
          \STATE \textbf{3. Observe:} Execute $a_t$, get observation $o_t$.
          \STATE \textbf{4. Update Belief ($b_{t+1}$):}
          \STATE \quad Extract new facts from $o_t$.
          \STATE \quad Mark satisfied predicates \emph{True}.
          \STATE \quad Append new sub-questions.
        \ENDWHILE
        \STATE \textbf{return} Best answer $a_{\text{final}}$ using $b_t$
      \end{algorithmic}
    \end{algorithm}
  \end{minipage}
\end{figure}

To build a reliable agent for the CGDP, we define the operations that should be performed. 
In this section, we introduce Predicate-Based Adaptive Identification (PBAI), an explicit algorithm template
for a CGDP agent loop. By mapping state-of-the-art agent harnesses to the PBAI operations, we can diagnose
where their implicit approximations fall short of desired behavior. 

\subsection{Predicate-Based Adaptive Identification}

In a CGDP, the full hidden state $c$ is too large for an LLM to hold in its context. 
To successfully navigate the CGDP, the agent's belief state $b_t$ must compress its understanding of the hidden state into a finite discrete object that can fit compactly in context.

We define a \textbf{predicate} as a logical proposition relevant to the task $q$ that can be evaluated
as \emph{True} or \emph{False} by querying the environment (e.g. ``The target function returns a list''in a codebase $c$). A useful belief state is one that tracks which predicates have been resolved by observed evidence and which remain unresolved. 
To maintain and act upon this state, the agent iterates through a four step loop that we call Predicate-Based Adaptive Identification (PBAI):
\begin{enumerate}[leftmargin=*,nosep]
\item \textbf{Stop?} The agent evaluates its belief state $b_t$. If the necessary predicates relevant to $q$ are resolved unambiguously, the agent outputs a final answer and terminates.
\item \textbf{Select Action} If unresolved predicates remain, the agent formulates a new hypothesis about the hidden state. It selects an action $a_t$ to resolve the highest-priority unresolved predicate.
\item \textbf{Observe} The agent executes $a_t$ against the environment $F(\cdot, c)$ and receives observation $o_t$.
\item \textbf{Update Belief} The agent updates its belief state $b_t$ given $o_t$. Belief updates can revise beliefs about the final answer, eliminate hypotheses inconsistent with $o_t$, update the agent’s understanding of the observation function $F(\cdot, c)$, or introduce new predicates to be resolved. 
\end{enumerate}

Algorithm~\ref{alg:ideal} provides pseudocode for the PBAI loop. While POMDP solvers maintain probability distributions over states, LLMs instead autoregressively generate a textual action string based on their prompt. 
PBAI abstracts this textual generation in the \textbf{Select Action} step as a form of approximate Thompson Sampling~\citep{russo2018tutorial}. While our experiments utilize greedy decoding (temperature $0$) for reproducibility, conceptually, the generation of a specific search query corresponds to the agent traversing its internal hypothesis space, seeking evidence to test its predicates, and updating its internal state. 

\subsection{Mapping Agent Harnesses via PBAI}

Existing Retrieval-Augmented Generation (RAG) and agentic memory methods perform all four PBAI operations to some extent, but they do so implicitly. 
By viewing these harnesses through PBAI, we can identify where they fall short (Table~\ref{tab:method-instantiation}).

\begin{table}[ht]
\centering
\caption{Four agent harnesses mapped to the PBAI algorithm. $\dagger$ denotes operations where infrastructure gaps can arise. Our experiments in Section~\ref{sec:experiments} confirm these gaps.}
\label{tab:method-instantiation}
\small
\renewcommand{\arraystretch}{1.2}
\begin{tabularx}{\textwidth}{@{}l X X X X@{}}
\toprule
 & \textbf{IRCoT} & \textbf{ReAct} & \textbf{Iter-RetGen} & \textbf{MemGPT} \\
\midrule
\textbf{Belief State} & Unbounded CoT \& retrieved passages$^\dagger$ & Unbounded trajectory history$^\dagger$ & None (memoryless)$^\dagger$ & LLM-managed memory block$^\dagger$ \\
\textbf{Step 1 (Stop?)} & Implicit (token generation)$^\dagger$ & Explicit \texttt{Finish} action & None (fixed rounds)$^\dagger$ & LLM-managed tool call$^\dagger$ \\
\textbf{Step 2 (Act)} & Implicit via next CoT & Explicit \texttt{Search} action & Previous generation$^\dagger$ & Explicit tool call \\
\textbf{Step 4 (Update)}& Implicit (append-only)$^\dagger$ & Implicit (append-only)$^\dagger$ & None$^\dagger$ & LLM-managed memory edit$^\dagger$ \\
\bottomrule
\end{tabularx}
\end{table}

\paragraph{IRCoT (implicit belief state)} The concatenation of all past CoT sentences and retrieved passages serves as IRCoT's belief state. While the CoT sentences act as an implicit \textbf{Update Belief} step, it is not curated or explicitly maintained. As the trajectory grows, earlier CoT sentences fall out of the LLM context, causing lossy representation and goal drift over long horizons.

\paragraph{ReAct (inefficient exploration)} The explicit \emph{Finish[answer]} action is available to an LLM agent to execute the \textbf{Stop?} step. However, 
ReAct agents frequently suffer from goal displacement, because they lack a dedicated mechanism to anchor unresolved predicates of the original user query $q$.

\paragraph{MemGPT (unreliable metacognition)} Explicit memory management tools enable an explicit \text{Update Belief} step. However, the \textbf{Stop?} step is left entirely to the LLM's self-assessment. Relying on weak metacognition leads to high rates of premature stopping or infinite tool-calling loops.

\paragraph{Iter-RetGen (memoryless)} Without a persistent belief state, no adaptive stopping (executing for a fixed number of rounds), and no belief update mechanism, this is the crudest PBAI approximation. 

In summary, current agent harnesses ask an LLM to act as the belief state, the CGDP policy, and the metacognitive evaluator simultaneously. To solve CGDP more reliably, we propose that PBAI operations be unbundled and have dedicated infrastructure for specific operations. 

\section{Interventions}\label{sec:fixes}

In this section, we derive two modular, orchestrator-level interventions that explicitly implement the PBAI operations. Because these interventions exist outside the agent's LLM prompt, they can be inserted into standard harnesses without difficulty. 

\subsection{Predicate-Based Belief State}\label{sec:fix1}

The most severe limitation of standard agent harnesses is the unbounded accumulation of context. 
To resolve this, we introduce a predicate-based belief state and an explicit implementation of the PBAI \textbf{Update Belief} step. This belief state $b_t$ replaces the unbounded trajectory history with a tightly constrained persistent data structure. It consists of two conceptually distinct elements:
\begin{enumerate}[leftmargin=*,nosep]
\item Facts: A curated list of confirmed propositions (i.e., predicates resolved with evidence) extracted from past observations. 
\item Open Predicates: A queue of unresolved sub-questions that must be answered to satisfy $q$. 
\end{enumerate}

Rather than relying on an LLM to remember all it has read during an episode, the orchestrator actively manages the belief state using a modular \textbf{Extractor}. At timestep $t$, 
the orchestrator passes $(b_t,o_t)$ to a lightweight LLM extraction call ($< 500$ tokens). The extractor parses $o_t$, appends newly discovered facts, marks resolved predicates, and appends sub-questions if $o_t$ reveals missing context. To ensure a bounded context footprint, the extractor compresses older findings and curates the state down to $K \le 6$ items (see Appendix~\ref{app:reorganization_cost} for token overhead analysis of this step). This updated state $b_{t+1}$ is injected into the agent's prompt for the next timestep, discarding the raw $o_t$. In contrast to the na\"ive truncation in Equation~\ref{eq:naive_state}, $b_{t+1} = \text{Extract}(b_t, o_t)$ distills a compact representation while preserving sufficient information. 

To understand how an LLM processes explicit $b_t$, we experiment (in Section~\ref{sec:experiments}) with two different textualizations of the predicate-based belief state by changing the extractor's output format:
\begin{itemize}[leftmargin=*,nosep]
\item Structured object: A JSON schema containing key-value fields for facts and open predicates.
\item Freeform text: A natural language summary that the LLM can use as a scratchpad. 
\end{itemize}
We provide the orchestrator prompts, schema definitions, and all extractor details in Appendix~\ref{app:prompts}. 

\subsection{Exhaustion Gate}\label{sec:fix2}

A secondary failure mode in the CGDP is improper stopping (the first step in PBAI). 
Relying on an LLM's self-assessment leads to two extremes: premature stopping (halting before sufficient context is gathered due to, e.g. generating a plausible hallucination) or infinite looping (repeatedly issuing similar queries due to, e.g. mode collapse). 
To prevent this, we introduce the Exhaustion Gate. 

Instead of relying on the LLM's metacognition to realize that its search has stagnated, the orchestrator explicitly monitors the search loop. 
We later compare both a focused lightweight LLM call (analogous to the extractor) and programmatic heuristics, and find that programmatic heuristics are reliable and more performant. 
The programmatic gate tracks two quantities across the agent's trajectory:
\begin{enumerate}[leftmargin=*,nosep]
\item Action similarity: Measures the lexical overlap between the current action $a_t$ and recent actions, e.g. using $Jaccard$ similarity. For example, with retrieval actions we compute $Jaccard$ over the strings queried in the actions; this helps to detect looping behavior. 
\item Novelty: Measures the overlap between the current observation $o_t$ and previous observations, e.g. using $UPR$ (Unique Passage Rate). UPR is the percentage of newly retrieved chunks in $o_t$ that have not been seen in previous rounds. A low UPR indicates that the agent's recent actions are no longer surfacing novel text from the hidden state $c$, suggesting that the current search hypothesis has run dry. 
\end{enumerate}
At each timestep $t$, the orchestrator evaluates $Stagnated_t \coloneqq (Jaccard_t \ge \tau_J) \wedge (UPR_t \le \tau_U)$. If $Stagnated_t$ remains \emph{True} for $p$ consecutive rounds, the orchestrator interrupts the PBAI loop and forces the agent to give a final answer based on the textualization of the current belief state $b_t$. 
To ensure our gate is not overly sensitive to specific hyperparameters or chunking strategies, we evaluated $16$ different threshold configurations (both discrete and smooth exponential moving averages). 
The exact values ($\tau_J, \tau_U, p$) used in our experiments, along with our robustness sweep, are detailed in Appendix~\ref{app:gate-configs}, demonstrating that the programmatic gate provides stable gains across a wide range of settings. 

\section{Experiments and Analysis}\label{sec:experiments}

To validate the CGDP framework, we apply PBAI interventions to four agent harnesses: IRCoT~\citep{trivedi2023ircot}, ReAct~\citep{yao2023react}, MemGPT~\citep{packer2024memgptllmsoperatingsystems} and Iter-RetGen~\citep{shao2023enhancing}. We design experiments to answer two questions:
\begin{enumerate}[leftmargin=*,nosep]
\item Does explicit extraction of a predicate-based belief state prevent performance degradation over long horizons? 
\item Does the programmatic exhaustion gate save tokens and prevent inefficient search better than LLM self-assessment? 
\end{enumerate}

\subsection{Setup}
\paragraph{Datasets:} We evaluated three complex retrieval-augmented QA domains requiring sophisticated exploration: LoCoMo~\citep{maharana2024locomo} (conversational QA), MuSiQue~\citep{trivedi2022musique} (multi-hop reasoning) and SWE-QA-Pro~\citep{sweqapro2025} (code repository QA). 

\paragraph{Model and Metrics:} All agents share identical zero-shot prompts and retriever configurations. Retrieval combines BM25 and \texttt{all-MiniLM-L6-v2} embeddings matching published harness weights ($0.9$ for IRCoT, $0.5$ otherwise). 
Agents and orchestrators use \texttt{gpt-4o-mini} (temperature $0$). Performance is evaluated via LLM-as-a-Judge (\texttt{gpt-4o} temperature $0$), which measures Correctness, Completeness, penalizes Irrelevance, and normalizes to $0-100\%$. 
Cost is measured in total LLM API tokens. 
To ensure LLM judge is unbiased toward specific formatting, we verified standard string-matching metrics (Token F1, Exact Match, ROUGE, and SBERT). As detailed in Appendix~\ref{app:string-metrics}, these lexical metrics broadly agree with the rubric judge rankings, confirming that the accuracy gains are substantive.
We use paired $t$-tests with Holm--Bonferroni correction.

\subsection{Lobotomize-then-Replace Methodology}\label{sec:design}

To isolate the impact of our interventions, we evaluate each agent harness under four controlled memory conditions:
\begin{enumerate}[leftmargin=*,nosep]
    \item Baseline: The harness runs as-is, accumulating its standard observation trajectory. 
    \item Lobotomized: The harness' native memory is wiped at every timestep. It sees only the current observation $o_t$ and the original query $q$. This tests how dependent the harness is on the trajectory history.
    \item PBBS (Structured, $b_t^{struct}$): The lobotomized harness is injected with a JSON object that maintains up to $K=6$ key-value pairs of facts and open predicates. 
    \item PBBS (Freeform, $b_t^{free}$): The lobotomized harness is injected with a natural language paragraph, rewritten each timestep by the extractor to distill facts and open predicates. 
\end{enumerate}

\subsection{Studying the Predicate-Based Belief State}\label{sec:pbbs_results}

Stripping an agent's memory (Lobotomized) causes severe degradation (Table~\ref{tab:main_results}), particularly on MuSiQue. However, replacing native memory management with an explicit predicate-based belief state completely prevents this degradation (all $b_t^{free}$ improvements are statistically significant, $p < 0.05$). 
Crucially, by distilling trajectories and preventing early evidence from falling out of the context window, it frequently outperforms baselines (e.g. $+ 11.4\%$ on MuSiQue, $p < 0.001$). 

\paragraph{Freeform text outperforms Structured JSON.} We observed that textualizing the belief state as a natural language running summary or freeform scratchpad ($b_t^{free}$) consistently outperformed ($10$ in $12$ experimental settings) textualizing it with a rigid JSON schema ($b_t^{struct}$). 
To understand why freeform text wins, we analyze the extraction outputs. When forced into a rigid JSON schema, the LLM generates artifacts like 'no evidence' for open questions (occurring in up to $56\%$ of episodes, see Table~\ref{tab:noevidence}). This artificially fragments the LLM's reasoning and misleadingly frames the state on answerable questions. Freeform text allows the LLM to organize the content in whatever natural language form best preserves the information.

\begin{table}[htbp]
\centering
\caption{Performance (Rubric Judge Accuracy, \%) across memory conditions. Adding the PBBS ($b_t$) to lobotomized agents fully recovers and often improves over the baselines. Bold indicates best per harness-task pair. Iter-RetGen (memoryless by design) has no lobotomized condition.}
\label{tab:main_results}
\small
\setlength{\tabcolsep}{4pt}
\begin{tabularx}{\textwidth}{@{}l X r r r r@{}}
\toprule
\textbf{Task} & \textbf{Harness} & \textbf{Baseline} & \textbf{Lobotomized} & \textbf{+ PBBS ($b_t^\text{struct}$)} & \textbf{+ PBBS ($b_t^\text{free}$)} \\
\midrule
\multirow{4}{*}{LoCoMo} 
 & IRCoT & 63.0 & 57.2 & 63.6 & \textbf{65.4} \\
 & ReAct & 65.1 & 59.4 & 65.7 & \textbf{66.6} \\
 & Iter-RetGen & \textbf{67.3} & -- & 65.0 & 65.9 \\
 & MemGPT & 57.2 & 56.5 & \textbf{65.4} & 62.5 \\
\midrule
\multirow{4}{*}{MuSiQue} 
 & IRCoT & 14.4 & 9.0 & 21.1 & \textbf{25.9} \\
 & ReAct & 29.6 & 12.2 & 30.1 & \textbf{36.8} \\
 & Iter-RetGen & 37.1 & -- & 33.8 & \textbf{39.6} \\
 & MemGPT & 31.4 & 30.6 & 29.6 & \textbf{34.4} \\
\midrule
\multirow{4}{*}{SWE-QA-Pro} 
 & IRCoT & 27.3 & 31.7 & 33.3 & \textbf{34.6} \\
 & ReAct & 39.8 & 33.7 & 37.3 & \textbf{43.0} \\
 & Iter-RetGen & 50.9 & -- & 48.4 & \textbf{51.4} \\
 & MemGPT & \textbf{48.9} & 43.6 & 42.3 & 47.7 \\
\bottomrule
\end{tabularx}
\end{table}

\begin{table}[!ht]
\centering
\caption{Prevalence of ``no evidence'' open questions in structured $b_t^{\text{struct}}$ across harnesses and tasks.}
\label{tab:noevidence}
\small
\begin{tabular}{ll cc}
\toprule
\textbf{Harness} & \textbf{Task} & \textbf{\% of open Qs} & \textbf{\% episodes affected} \\
\midrule
IRCoT & MuSiQue & 37.7\% & 63.9\% \\
IRCoT & LoCoMo & 56.5\% & 93.7\% \\
IRCoT & SWE-QA-Pro & 41.2\% & 72.3\% \\
ReAct & MuSiQue & 40.7\% & 72.6\% \\
ReAct & LoCoMo & 56.2\% & 97.0\% \\
ReAct & SWE-QA-Pro & 42.3\% & 76.9\% \\
\midrule
\multicolumn{2}{l}{Freeform (all)} & 0\% & 0\% \\
\bottomrule
\end{tabular}
\end{table}

\begin{figure}[htbp]
    \centering
    \includegraphics[width=\textwidth]{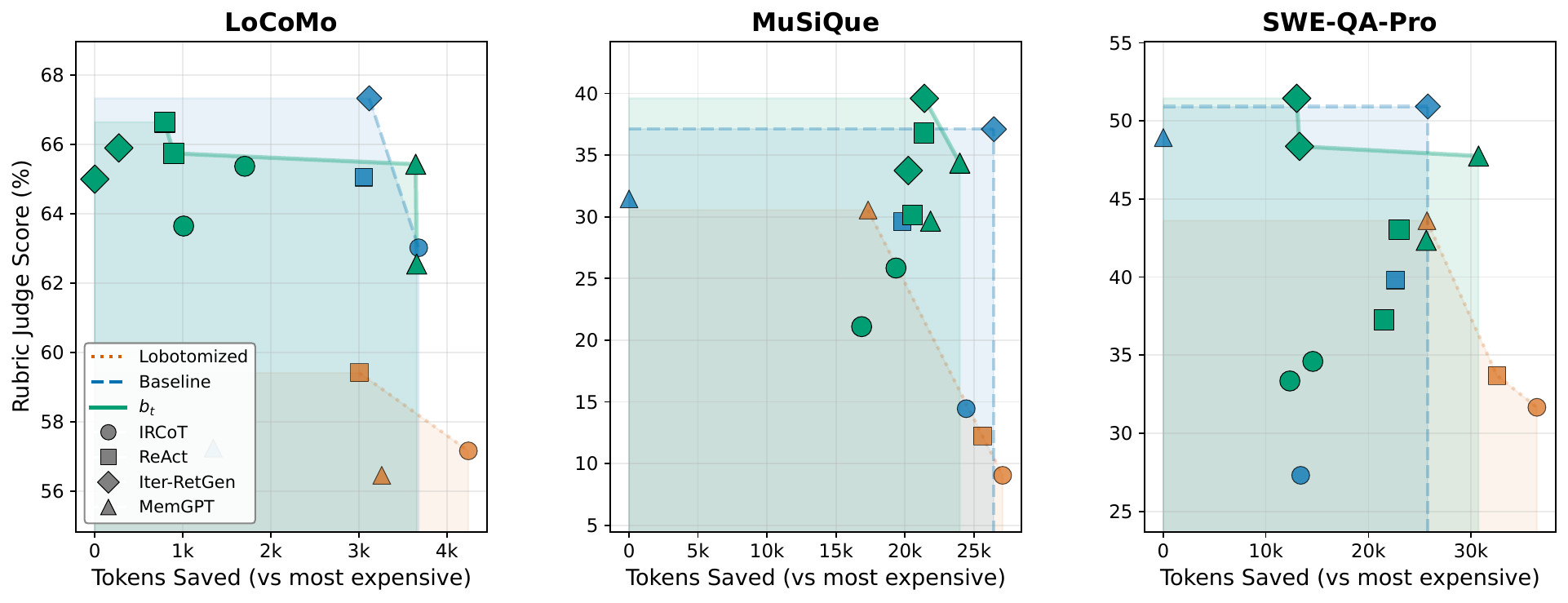}
    \caption{Pareto frontiers of quality vs.\ token efficiency across harnesses. Lobotomization (orange) collapses the frontier; $b_t$ (green) recovers much of it toward baseline levels (blue).}
    \label{fig:pareto}
\end{figure}

While adding orchestrator steps might intuitively seem to increase overall token costs, Figure~\ref{fig:pareto} shows that the belief state fundamentally improves the quality-cost trade-off. Lobotomization (orange) collapses the Pareto frontier; injecting the belief state (green) recovers the frontier toward baseline levels (blue), so the accuracy gains do not come at an increased token cost. 

\subsection{Studying the Programmatic Exhaustion Gate} \label{sec:gate_results}

As shown in Table~\ref{tab:gate_avg}, using the programmatic exhaustion gate on stateful baselines improves accuracy across all three domains (up to $+6.9\%$ on SWE-QA-Pro). 
Importantly, the gate relies on the belief state $b_t$ to generate the final answer; when applied to lobotomized agents (which do not have accumulated facts), stopping the loop early results in a slight accuracy drop. 

\paragraph{Programmatic heuristic outperforms LLM metacognition.} Table~\ref{tab:gate_llm_compare} compares the programmatic heuristic against prompting the LLM to self-assess stagnation. 
The LLM can be prompted in two distinct ways and shows two extreme failures. If prompted to act conservatively, it spirals into infinite loops (unfounded skepticism), costing $159\%$ more tokens than the baseline. If prompted neutrally, it triggers premature stopping that damages accuracy by $5.0\%$. In contrast, the programmatic gate does not require additional LLM calls and saves up to $39\%$ of the total tokens without sacrificing accuracy. 

\begin{table}[htbp]
\centering
\begin{minipage}[t]{0.48\textwidth}
\centering
\caption{Exhaustion Gate impact on task accuracy (averaged across IRCoT, ReAct, MemGPT). Per-harness results are in Appendix~\ref{app:gate-configs}. Bold indicates statistical significance ($p < 0.05$). 
}
\label{tab:gate_avg}
\small
\begin{tabular}{llrr}
\toprule
\textbf{Task} & \textbf{Condition} & \textbf{Base Acc.} & \textbf{$\Delta$ w/ Gate} \\
\midrule
LoCoMo & Baseline & 61.8\% & \textbf{+1.1\%} \\
MuSiQue & Baseline & 25.1\% & \textbf{+2.8\%} \\
SWE-QA & Baseline & 38.7\% & \textbf{+6.9\%} \\
\midrule
MuSiQue & Lobotomized & 17.3\% & $-2.9\%$ \\
\bottomrule
\end{tabular}
\end{minipage}\hfill
\begin{minipage}[t]{0.48\textwidth}
\centering
\caption{Programmatic vs. LLM-assessed exhaustion gates on IRCoT in SWE-QA-Pro. Programmatic saves tokens without degrading accuracy, LLM costs more tokens than it saves. Bold indicates statistical significance ($p < 0.05$). }
\label{tab:gate_llm_compare}
\small
\begin{tabular}{lrr}
\toprule
\textbf{Gate Type} & \textbf{Acc. $\Delta$} & \textbf{Token Savings} \\
\midrule
Programmatic & \textbf{+16.2\%} & \textbf{+39.0\%} \\
LLM (Conservative) & $-3.3\%$ & $-159.0\%$ \\
LLM (Neutral) & $-5.0\%$ & $+56.4\%$ \\
\bottomrule
\end{tabular}
\end{minipage}
\end{table}

\section{Discussion and Conclusion}\label{sec:discussion}

Formalizing iterative agentic search as a Context Gathering Decision Process (CGDP) provides a blueprint for designing reliable LLM harnesses. Viewing agent behavior through approximate Thompson Sampling, our framework is able to identify modular infrastructure interventions. We demonstrated that by explicitly unbundling the operations of the search loop, we can systematically mitigate catastrophic failures. The two interventions derived from this framework -- a persistent belief state and a programmatic exhaustion gate -- compose with each other in state-of-the-art harnesses. They are empirically effective in preventing context degradation and halting unproductive search, saving up to $39\%$ of tokens while improving multi-hop reasoning across three domains.

Our empirical evaluation surfaced a design principle: orchestrators should dictate \textit{operations} not \textit{representations}. While the orchestrator enforces state tracking steps (extraction, curation), imposing rigid structures (e.g. JSON) on intermediate representation interferes with LLM reasoning. The framework is most effective when it guides control flow while letting the LLM freely synthesize evidence in a natural language scratchpad. 

\paragraph{Limitations and Future Work.} All evaluations in this study use a single agent LLM (\texttt{GPT-4o-mini}). While our framework predicts even stronger gains on weaker models where implicit state tracking degrades faster, cross-model generalization remains an open question for future work. 
Furthermore, while the CGDP models general hidden states (e.g. codebases with BASH tools), our empirical validation focuses on complex, long-horizon retrieval to establish baseline efficacy. Finally, the CGDP serves as a conceptual and empirical framework to guide infrastructure design, rather than providing formal mathematical bounds. 

Ultimately, the CGDP abstraction is valuable for pinpointing infrastructure gaps before they cause silent failures in practice. 
Our analysis reveals several unaddressed gaps in current harnesses -- such as LLM priors misaligned with the observation function and error compounding across trajectories (detailed in Appendix~\ref{app:phenomena}). By mapping to the operations of the Predicate-Based Adaptive Identification loop, we can find fruitful ways to address each of them in future work. 

\begin{ack}
We thank Ambuj Tewari, the Netflix Machine Learning and Inference Research group, Ding Tong, Aditya Sinha, Maya Ravichandran, and Anuj Phadke for their insightful feedback on this project. 
\end{ack}

\bibliographystyle{plainnat}
\bibliography{refs}


\appendix


\section{Observed Failure Modes and Example Traces}\label{app:phenomena}

\subsection{Observed Failure Modes}
This section summarizes the main failure modes (G1–G7) surfaced by our evaluation tasks, together with several additional failure modes that the PBAI framework suggests may arise in longer episodes, larger corpora, or harder reasoning settings. 

\paragraph{Coverage unawareness.} The agent cannot tell the difference between ``I did not find it'' and ``it is not in the corpus''. Current methods usually have little notion of corpus coverage, so they cannot estimate how much of the search space has already been explored. A natural fix would be explicit coverage tracking.

\paragraph{Aggregation breakdown.} Some questions require combining many weak signals across passages, such as counting entities, summarizing trends, or comparing sources. LLMs become less reliable as the number of relevant passages grows~\citep{schnabel2025losttransmission}, so the state update can fail even when the evidence has already been retrieved.

\paragraph{Error compounding.} Small extraction mistakes in early rounds can become premises for later reasoning. Persistent state helps by keeping each extraction step short and allowing later revisions, but it does not fully remove this problem.

\paragraph{Action-observation misalignment.} The agent may search the wrong part of the corpus because its internal picture of how actions map to observations is wrong. In practice, this means it chooses actions that are locally sensible but poorly matched to the corpus structure. A natural fix would be a lightweight predict-before-retrieve step.

\paragraph{Unjustified confidence.} The agent may stop with high confidence because the answer sounds plausible, not because the evidence is complete. This happens when stopping depends more on fluent output than on whether the key predicates have actually been resolved. A natural fix would be evidence-grounded stopping or calibrated answer generation.

\subsection{Annotated Agent Traces}\label{app:traces}
All traces below are from baseline conditions of our experiments (no $b_t$ injection, no exhaustion gate).

\paragraph{G5/G6: Retrieval Stagnation (IRCoT, MuSiQue).} Retrieval stagnation occurs when the agent issues the same or near-identical search actions across rounds, receiving the same passages and making no progress. We measure this as episodes where the token-level Jaccard similarity between consecutive actions exceeds 0.5 for at least 2 consecutive pairs.\\
\textbf{Example:} User query: \emph{``What city shares a border with the place where the person who went to the state known for its Mediterranean climate during the gold rush worked?''} Ground truth: Rio Linda. IRCoT runs 10 rounds (maximum) and answers incorrectly.
\begin{quote}\small
R1: \texttt{To answer the question, we need to identify a city that shares a border with a location associated with the Mediterranea...} \\
R2: \texttt{The Black Hills Gold Rush took place in Dakota Territory, which is not associated with a Mediterranean climate.} \\
R3: \texttt{To answer the question, we need to identify a city that shares a border...} \\
R5--R10: Repeat of R1/R3 pattern with minor variations.
\end{quote}
\textbf{Annotation.} The agent alternates between two framings without advancing the multi-hop chain. Because it has no explicit mechanism for detecting stagnation or trying a new direction, it remains stuck in the loop. The exhaustion gate would flag this quickly via high action overlap. 

\paragraph{G4: Premature Stopping (MemGPT, LoCoMo).} Premature stopping occurs when the agent stops after 1--2 rounds with an incorrect answer despite insufficient evidence.\\
\textbf{Example.} User query: \emph{``Who performed at the concert at Melanie's daughter's birthday?''} Ground truth: Matt Patterson. MemGPT runs 1 round, retrieves 5 passages with action \texttt{``concert Melanie's daughter's birthday''}, and answers \textsc{UNANSWERABLE}.\\
\textbf{Annotation.} The agent does not find the answer in the first retrieval and stops instead of refining the search. With a persistent belief state, the unresolved question ``who performed?'' would remain explicit, making early termination less attractive.

\paragraph{G2: Goal Displacement (ReAct, MuSiQue).} Goal displacement occurs when the agent's actions drift from the original user query, pursuing tangential leads.\\
\textbf{Example.} User query: \emph{``How were people from whom new coins were a proclamation of independence by the Somali Muslim Ajuran Empire expelled from the natural boundary between Thailand and A Magne's country?''} Ground truth: The dynasty regrouped and defeated the Portuguese. ReAct runs 5 rounds.
\begin{quote}\small
R1 action: \texttt{Ajuran Empire proclamation of independence and expulsion of people} \\
R5 action: \texttt{Ajuran Empire Southeast Asia history}
\end{quote}
\textbf{Annotation.} The agent follows a salient thread but drifts away from the intermediate sub-questions required to answer the original query. Without explicit state tracking for what remains unresolved, tangential directions are easy to pursue.

\paragraph{G1: Evidence Not Persisted (IRCoT, MuSiQue).} Evidence forgetting occurs when relevant information appears in early rounds but is absent from later reasoning.\\
\textbf{Example.} User query: \emph{``Who is the mother of the screenwriter of WarGames?''} Ground truth: Jane Greer. IRCoT runs 10 rounds and answers incorrectly. Round 2 passage contains: \emph{``...He is the son of actress Jane Greer and producer Edward Lasker...''} By rounds 8--10, the chain-of-thought discusses the screenwriters without mentioning their mothers.\\
\textbf{Annotation.} The relevant evidence appears early but is not preserved in a durable form, so it no longer shapes later reasoning. A good persistent belief state would have written ``Jane Greer'' into the running state as soon as it appeared.

\paragraph{G3: Parametric Dominance (ReAct, MuSiQue).} Parametric dominance occurs when the agent produces an answer grounded in pretraining knowledge rather than the retrieved evidence.\\
\textbf{Example.} User query: \emph{``How many mandatory transmitters of the Canadian Broadcasting Centre's owner were updated before the deadline?''} Ground truth: only about half. ReAct runs 3 rounds.
\begin{quote}\small
R2 Thought: \texttt{The observations indicate that the CBC did not convert all of its mandatory transmitters to digital by the original deadline of August 31, 2011...} \\
Answer: \texttt{15 mandatory transmitters were updated before the deadline}
\end{quote}
\textbf{Annotation.} The final answer introduces a specific number that is not supported by the retrieved evidence. Here the model’s parametric prior wins over the retrieved passages because the evidence is not kept in an explicit, persistent state.

\paragraph{G7: Evidence-Action Misalignment (LoCoMo).} Agents incorporate retrieved evidence without verifying that it actually pertains to the specific entities in the user query.\\
\textbf{Example.} User query: \emph{``What was grandma's gift to Melanie?''} (entity-swapped; the original conversation discusses a gift to a different person). IRCoT retrieves passages about the necklace gift and answers ``A necklace'' in 2 rounds without verifying that the evidence pertains to Melanie specifically.\\
\textbf{Annotation.} The agent finds a relevant gift passage, but it does not verify that the gift is attached to the correct entity in the user query. Entity-grounded extraction or a lightweight alignment check would catch this mismatch before answer generation.

\begin{table}[!ht]
\centering
\caption{Gap prevalence across baseline conditions. Each cell shows the count of episodes exhibiting the pattern.}
\label{tab:trace-stats}
\small
\begin{tabular}{ll rrrr}
\toprule
\textbf{Gap} & \textbf{Task} & \textbf{IRCoT} & \textbf{ReAct} & \textbf{MemGPT} & \textbf{Iter-RetGen} \\
\midrule
\multirow{3}{*}{G4 (premature stop)} & MuSiQue & 74 & 4 & 62 & 0 \\
 & LoCoMo & 68 & 10 & 143 & 0 \\
 & SWE-QA-Pro & 15 & 0 & 3 & 0 \\
\midrule
\multirow{3}{*}{G5/G6 (stagnation)} & MuSiQue & 115 & 146 & 173 & 139 \\
 & LoCoMo & 248 & 70 & 24 & 986 \\
 & SWE-QA-Pro & 197 & 51 & 71 & 37 \\
\midrule
\multirow{3}{*}{G2 (goal displacement)} & MuSiQue & 114 & 236 & 180 & 0 \\
 & LoCoMo & 76 & 54 & 6 & 0 \\
 & SWE-QA-Pro & 9 & 61 & 20 & 0 \\
\midrule
G7 (evidence-action misalign.) & LoCoMo una. & 125/440 & 73/440 & 229/440 & --- \\
\bottomrule
\end{tabular}
\end{table}

\section{Additional Interventions Suggested by the Framework}\label{app:further-fixes}

The gap analysis suggests several additional orchestrator-level interventions that are outside the scope of this paper but follow naturally from the framework:

\paragraph{Gap-aware stopping.} Instead of asking only whether search has stalled, the orchestrator could prompt a focused LLM call to identify which predicates remain unresolved. This would directly target premature stopping and would complement the exhaustion gate, which targets the opposite failure mode of stagnation.

\paragraph{Entity-grounded extraction and alignment checking.} The extractor could be required to ground each finding against the entities and relations in the user query, and a separate verification step could catch remaining mismatches before answer generation.

\paragraph{Triggered resampling.} When stagnation is detected, the orchestrator could branch to one or more new hypotheses that are still consistent with the accumulated observations, rather than terminating immediately. This would make direction changes explicit rather than leaving them to the model’s chain-of-thought.

\paragraph{Observation filtering.} Before extraction, the orchestrator could filter or compress the raw observation itself, removing clearly irrelevant text before it ever reaches the belief-update step. 

\paragraph{Parallel hypothesis aggregation.} Another way to explore multiple hypotheses is to run several retrieval trajectories in parallel and aggregate them afterward~\citep{lee2026agenticaggregation}. This turns hypothesis diversity into an explicit orchestrator decision rather than an accidental byproduct of a single trajectory. 

\section{Extended Related Work}\label{app:extended-related}

\subsection{Agentic RAG Landscape}
Recent work has explored giving retrieval agents more sophisticated decision-making capabilities. SearchR1~\citep{jin2025searchr1} and Search-o1~\citep{li2025searcho1} train retrieval agents with reinforcement learning. WebThinker~\citep{li2025webthinker} extends this to web-scale retrieval. SLIM~\citep{yen2025lostmaze} targets long-horizon search by periodically summarizing trajectories. These approaches are complementary to ours: they improve the agent's action selection policy or trajectory management through training and tool design, while we improve the infrastructure around the policy (persistent state, exhaustion detection) through orchestration.

\subsection{LLM Metacognition and Self-Assessment}
Our finding that programmatic exhaustion detection is more token-efficient than LLM-judged stopping is consistent with a growing body of evidence on LLM metacognitive limitations. \citet{feng2024abstain} showed that multi-LLM probing improves abstention by 19.3\% over single-model self-assessment. \citet{guo2024calibration} demonstrated that verbalized confidence is poorly calibrated across model families. These findings motivate our design choice: the stagnation signals we measure (action Jaccard and UPR) are observable facts about the retrieval process that do not require the LLM to assess its own search progress.

\subsection{Consistency and Contradiction Detection}
Evidence-action misalignment connects to a literature on contradiction detection in LLM outputs. \citet{mundler2024selfcontradictory} showed that pairwise contradiction detection achieves approximately 80\% F1, substantially better than document-level detection. The key insight is that contradiction detection is effective when the relevant facts are presented together in short context, but degrades when they must be identified from a long history. This motivates the orchestrator's role: maintaining a persistent state ($b_t$) so that new extractions can be compared against existing facts in focused pairwise calls.

\subsection{Generative Agents and Long-Term Memory}
Park et al.~\citep{park2023generativeagents} introduced generative agents with long-term memory for simulated social environments. Their memory architecture (observation $\to$ reflection $\to$ planning) shares structural similarities with our belief state management (observation $\to$ extraction $\to$ state update), but operates in a different setting: their agents interact with a changing social world (full POMDP), while our agents search a static database (sequential identification). The MAKER framework~\citep{meyerson2025maker} extends agentic memory to long-horizon creative tasks.


\section{Experimental Setup and Reproducibility}\label{app:setup}

\paragraph{Datasets and Splits.} We evaluated three datasets: LoCoMo (1,275 tasks spanning conversational memory), MuSiQue (500 answerable tasks spanning multi-hop Wikipedia routing), and SWE-QA-Pro (260 tasks spanning software repository structures). We utilized the standard test/validation splits provided by the authors of the respective benchmarks.

\paragraph{Compute Resources and LLM Usage.} All agent trajectories, modular extractions, and rubric evaluations were executed via OpenAI's API using the \texttt{gpt-4o-mini} and \texttt{gpt-4o} endpoints. The total compute expenditure for the experiments, including baseline runs, lobotomization sweeps, and gate ablations, was approximately $\$1,500$ in API credits.

\paragraph{Licenses.} We use LoCoMo (\textbf{CC BY-NC 4.0}), MuSiQue (\textbf{CC BY 4.0}), and SWE-QA-Pro Bench (\textbf{MIT}) as evaluation datasets, and \texttt{all-MiniLM-L6-v2} (\textbf{Apache-2.0}) for dense retrieval embeddings. We accessed \texttt{gpt-4o-mini} and \texttt{gpt-4o} through the OpenAI API under the applicable OpenAI Services Agreement and Service Terms. 

\section{Orchestrator and Judge Prompts}\label{app:prompts}

We list the key prompts designed for our interventions. All prompts are in the experiment codebase under \texttt{src/orchestrator/}. Harness-specific prompts and rubric judge prompts are in \texttt{src/methods/prompts/} and \texttt{src/scoring/}.

\subsection{Structured Extraction Prompt}
The structured prompt separates output into facts, resolved questions, and new questions. The instruction to ``state precisely what evidence is still missing'' produces the ``no evidence'' artifacts analyzed in Section~6.

\begin{quote}\small\ttfamily
New retrieved passages:\\
\{observation\}\\[4pt]
What we already know (DO NOT repeat any of these):\\
\{established\_facts\}\\[4pt]
<scratchpad>\\
First, what do the passages actually state? Extract the key claims exactly as the evidence presents them --- preserve who did what, which entity is involved, what values are mentioned. Do not paraphrase in a way that changes attribution or meaning.\\
Then, given the question "\{question\}", what is new here compared to what we already know? Do any of these claims resolve our open questions? If a fact is already listed above, skip it entirely.\\
</scratchpad>\\[4pt]
Open questions we are still investigating:\\
\{open\_questions\}\\[4pt]
Output ONLY genuinely new facts not already listed above. If the passages contain nothing new beyond what we already know, write "Nothing relevant."\\[4pt]
New facts:\\
- The claim exactly as the evidence states it (source: document or passage identifier)\\[2pt]
Resolved questions:\\
- Which open questions are now answered by the evidence?\\[2pt]
New questions:\\
- What specific evidence is still missing? State precisely what has NOT been found.
\end{quote}

\subsection{Freeform Extraction Prompt}
The freeform prompt produces notes and memories without fact/question separation. 

\begin{quote}\small\ttfamily
New retrieved passages:\\
\{observation\}\\[4pt]
Your current notes (DO NOT repeat any of these):\\
\{existing\_notes\}\\[4pt]
<scratchpad>\\
First, what do the passages actually state? Extract claims exactly as presented --- preserve who did what, which entities are involved. Do not paraphrase in a way that changes attribution.\\
Then, how do these relate to the question "\{question\}" and your current notes? What is genuinely new? If a note already covers this information, skip it entirely.\\
</scratchpad>\\[4pt]
Write ONLY genuinely new notes not already covered above. If the passages contain nothing new beyond your existing notes, write "Nothing relevant."\\[4pt]
Notes:\\
- A finding exactly as the evidence presents it\\[2pt]
Memories to keep:\\
- Verbatim quote or key passage worth preserving
\end{quote}

Both prompts operate on $b_t + o_t$ only (approximately 500 tokens of context), not the full history. 

\subsection{Reorganization Prompt}
To strictly bound the size of the belief state, the orchestrator allows the state to grow up to $K_{\text{trigger}} = 10$ items before pausing to execute a reorganization call. This call curates the state back down to a target size of $K_{\text{target}} = 6$ (the capacity limit reported in the main text).

\begin{quote}\small\ttfamily
You are curating an investigation state. Given the original question and all facts and questions gathered so far, produce a compact, prioritized state.\\[4pt]
Original question: \{question\}\\[4pt]
Current facts: \{facts\}\\[4pt]
Open questions: \{questions\}\\[4pt]
Instructions:\\
- Keep at most \{k\_target\} facts and \{n\_questions\} open questions\\
- Merge redundant facts into single comprehensive claims\\
- Drop facts irrelevant to the question\\
- PRESERVE facts that form multi-hop reasoning chains, even if individually they seem tangential\\
- Each fact must retain its source attribution\\
- Rewrite open questions to reflect what is actually still unknown\\
- Remove questions that have been answered by the facts\\
- Order facts by importance (most important first)
\end{quote}

\subsection{LLM-based Exhaustion Gate Prompts}
Two LLM-based stagnation detection variants were evaluated against the programmatic gate. Both see the current investigation state and recent retrieval rounds.

\paragraph{Conservative (v3).} Defaults to CONTINUE; requires concrete evidence of stagnation to recommend stopping.

\begin{quote}\small\ttfamily
You are deciding whether further retrieval will meaningfully improve the answer to this question.\\[4pt]
QUESTION: \{question\}\\[4pt]
CURRENT INVESTIGATION STATE: \{current\_state\}\\[4pt]
RECENT RETRIEVAL (last \{window\} rounds): \{recent\_rounds\}\\[4pt]
Your DEFAULT is to CONTINUE retrieval. Only recommend stopping if you can point to CONCRETE evidence of stagnation:\\
- The same passages or near-paraphrases are appearing across multiple rounds\\
- Search queries have covered the obvious angles and a meaningfully different direction is hard to identify\\
- The current answer already addresses the question and further evidence is unlikely to change it\\[4pt]
Do NOT recommend stopping just because:\\
- The answer seems plausible (it may still be incomplete)\\
- A few passages overlap (some overlap is normal)\\
- You are uncertain about the answer quality (uncertainty means more retrieval could help)\\[4pt]
VERDICT: PRODUCTIVE / QUERY\_STALE / EXHAUSTED\\
REASON: [explanation]
\end{quote}

\paragraph{Neutral (v3\_neutral).} No default direction; presents the three verdicts symmetrically.

\begin{quote}\small\ttfamily
You are evaluating whether an information retrieval investigation is making progress or has stagnated.\\[4pt]
QUESTION: \{question\}\\[4pt]
CURRENT INVESTIGATION STATE: \{current\_state\}\\[4pt]
RECENT RETRIEVAL (last \{window\} rounds): \{recent\_rounds\}\\[4pt]
Based on the evidence above, choose one of the following verdicts:\\[4pt]
VERDICT: PRODUCTIVE --- if new, relevant information is still being discovered each round.\\
VERDICT: QUERY\_STALE --- if the current search direction is exhausted but a specific untried angle could yield new information.\\
VERDICT: EXHAUSTED --- if retrieval has stalled and further rounds are unlikely to surface new relevant content.\\[4pt]
VERDICT: \\
REASON:
\end{quote}


\section{Full Results and Additional Metrics}\label{app:full-results}

\subsection{Rubric Judge Scores and Statistical Tests}
Tables~\ref{tab:full-ttests} and \ref{tab:locomo-ansuna} provide the comprehensive paired $t$-tests and split-condition scores that support the summarized findings in Section~6 of the main text.

\begin{table}[!ht]
\centering
\caption{Full paired $t$-tests for persistent belief state ($b_t$). Each cell reports the left condition minus the right condition in pp; positive means the left condition scores higher. Iter-RetGen has no lobotomized condition. {\color[rgb]{0,0,0.6}\textbf{Bold blue}} = $p < .05$.}
\label{tab:full-ttests}
\footnotesize
\setlength{\tabcolsep}{3pt}
\begin{tabular}{ll rrrrrr}
\toprule
 & & \multicolumn{3}{c}{\textbf{Lobotomization cost}} & \multicolumn{2}{c}{\textbf{$b_t$ recovery}} & \textbf{Format} \\
\cmidrule(lr){3-5} \cmidrule(lr){6-7} \cmidrule(lr){8-8}
\textbf{Task} & \textbf{Harness} & \textbf{B$-$L} & \textbf{B$-b_t^{\text{s}}$} & \textbf{B$-b_t^{\text{f}}$} & \textbf{L$-b_t^{\text{s}}$} & \textbf{L$-b_t^{\text{f}}$} & $\mathbf{b_t^{\text{s}}{-}b_t^{\text{f}}}$ \\
\midrule
\multirow{4}{*}{LoCoMo} & IRCoT & $\mathbf{\color[rgb]{0,0,0.6}+5.9}$ & $-0.6$ & $\mathbf{\color[rgb]{0,0,0.6}-2.4}$ & $\mathbf{\color[rgb]{0,0,0.6}-6.5}$ & $\mathbf{\color[rgb]{0,0,0.6}-8.2}$ & $-1.7$ \\
 & ReAct & $\mathbf{\color[rgb]{0,0,0.6}+5.6}$ & $-0.7$ & $-1.6$ & $\mathbf{\color[rgb]{0,0,0.6}-6.3}$ & $\mathbf{\color[rgb]{0,0,0.6}-7.2}$ & $-0.9$ \\
 & Iter-RetGen & --- & $\mathbf{\color[rgb]{0,0,0.6}+2.3}$ & $+1.4$ & --- & --- & $-0.9$ \\
 & MemGPT & $+0.8$ & $\mathbf{\color[rgb]{0,0,0.6}-8.2}$ & $\mathbf{\color[rgb]{0,0,0.6}-5.3}$ & $\mathbf{\color[rgb]{0,0,0.6}-9.0}$ & $\mathbf{\color[rgb]{0,0,0.6}-6.1}$ & $\mathbf{\color[rgb]{0,0,0.6}+2.9}$ \\
\midrule
\multirow{4}{*}{MuSiQue} & IRCoT & $\mathbf{\color[rgb]{0,0,0.6}+5.4}$ & $\mathbf{\color[rgb]{0,0,0.6}-6.7}$ & $\mathbf{\color[rgb]{0,0,0.6}-11.4}$ & $\mathbf{\color[rgb]{0,0,0.6}-12.0}$ & $\mathbf{\color[rgb]{0,0,0.6}-16.8}$ & $\mathbf{\color[rgb]{0,0,0.6}-4.8}$ \\
 & ReAct & $\mathbf{\color[rgb]{0,0,0.6}+17.3}$ & $-0.6$ & $\mathbf{\color[rgb]{0,0,0.6}-7.2}$ & $\mathbf{\color[rgb]{0,0,0.6}-17.9}$ & $\mathbf{\color[rgb]{0,0,0.6}-24.6}$ & $\mathbf{\color[rgb]{0,0,0.6}-6.7}$ \\
 & Iter-RetGen & --- & $+3.3$ & $-2.5$ & --- & --- & $\mathbf{\color[rgb]{0,0,0.6}-5.8}$ \\
 & MemGPT & $+0.9$ & $+1.8$ & $-2.9$ & $+0.9$ & $\mathbf{\color[rgb]{0,0,0.6}-3.8}$ & $\mathbf{\color[rgb]{0,0,0.6}-4.7}$ \\
\midrule
\multirow{4}{*}{SWE-QA} & IRCoT & $\mathbf{\color[rgb]{0,0,0.6}-4.4}$ & $\mathbf{\color[rgb]{0,0,0.6}-6.0}$ & $\mathbf{\color[rgb]{0,0,0.6}-7.3}$ & $-1.7$ & $\mathbf{\color[rgb]{0,0,0.6}-2.9}$ & $-1.3$ \\
 & ReAct & $\mathbf{\color[rgb]{0,0,0.6}+6.1}$ & $+2.6$ & $\mathbf{\color[rgb]{0,0,0.6}-3.2}$ & $\mathbf{\color[rgb]{0,0,0.6}-3.6}$ & $\mathbf{\color[rgb]{0,0,0.6}-9.3}$ & $\mathbf{\color[rgb]{0,0,0.6}-5.8}$ \\
 & Iter-RetGen & --- & $\mathbf{\color[rgb]{0,0,0.6}+2.6}$ & $-0.5$ & --- & --- & $\mathbf{\color[rgb]{0,0,0.6}-3.1}$ \\
 & MemGPT & $\mathbf{\color[rgb]{0,0,0.6}+5.3}$ & $\mathbf{\color[rgb]{0,0,0.6}+6.6}$ & $+1.2$ & $+1.3$ & $\mathbf{\color[rgb]{0,0,0.6}-4.1}$ & $\mathbf{\color[rgb]{0,0,0.6}-5.4}$ \\
\midrule
\multicolumn{2}{l}{Sig.\ pos / neg} & 7 / 1 & 3 / 3 & 0 / 6 & 0 / 6 & 0 / 9 & 1 / 7 \\
\bottomrule
\end{tabular}
\end{table}

\begin{table}[!ht]
\centering
\caption{LoCoMo $b_t$ effect (pp) split by answerable (835q) and unanswerable (440q entity-swap). Each cell reports the $b_t$ condition minus the comparison condition, so positive means $b_t$ scores higher. Structured $b_t$ improves unanswerable detection (abstention) while freeform $b_t$ improves answerable accuracy. Iter-RetGen has no lobotomized condition.}
\label{tab:locomo-ansuna}
\footnotesize
\setlength{\tabcolsep}{3pt}
\begin{tabular}{l rrrr rrrr}
\toprule
 & \multicolumn{4}{c}{\textbf{Answerable (835q)}} & \multicolumn{4}{c}{\textbf{Unanswerable (440q)}} \\
\cmidrule(lr){2-5} \cmidrule(lr){6-9}
\textbf{Harness} & $\mathbf{b_t^s{-}B}$ & $\mathbf{b_t^f{-}B}$ & $\mathbf{b_t^s{-}L}$ & $\mathbf{b_t^f{-}L}$ & $\mathbf{b_t^s{-}B}$ & $\mathbf{b_t^f{-}B}$ & $\mathbf{b_t^s{-}L}$ & $\mathbf{b_t^f{-}L}$ \\
\midrule
IRCoT & +1.0 & +5.8 & +12.1 & +16.9 & $-$0.1 & $-$4.0 & $-$4.2 & $-$8.1 \\
ReAct & $-$1.4 & +5.8 & $-$4.0 & +3.2 & +4.6 & $-$6.4 & +25.8 & +14.8 \\
Iter-RetGen & $-$8.0 & +0.1 & --- & --- & +8.3 & $-$4.4 & --- & --- \\
MemGPT & $-$5.0 & $-$1.6 & $-$5.3 & $-$1.9 & +33.2 & +18.4 & +36.1 & +21.3 \\
\bottomrule
\end{tabular}
\end{table}

\subsection{Lexical and Embedding-based Metrics}\label{app:string-metrics}

String metrics (Token F1, Exact Match, ROUGE-1, METEOR, SentenceBERT cosine similarity) broadly agree with the rubric judge rankings, confirming that the performance gains are not driven by judge preference for $b_t$-conditioned answers. On MuSiQue, $b_t^{\text{free}}$ is the best condition for every method on every string metric. 

The few divergences between string metrics and the rubric judge justify our primary use of the rubric judge (Table~\ref{tab:full-ttests}). For example, perfectly correct abstentions (e.g., answering "UNANSWERABLE") have zero token overlap with gold answers, artificially deflating F1 scores on LoCoMo despite being the optimal agent behavior. Similarly, SWE-QA-Pro string metrics favor lobotomized models because longer, verbose code explanations coincidentally have higher token overlap with the ground truth. The rubric judge accurately captures these task-specific quality dimensions (abstention, correctness, entity attribution) that surface-level overlap cannot.

\begin{table}[!ht]
\centering
\caption{Token F1 and Exact Match: $b_t$ effect (pp). Iter-RetGen (memoryless by design) has no lobotomized condition. {\color[rgb]{0,0,0.6}\textbf{Bold blue}} = $p < .05$.}
\label{tab:string-f1-em}
\small
\setlength{\tabcolsep}{3pt}
\begin{tabular}{ll rrrr rrrr}
\toprule
 & & \multicolumn{4}{c}{\textbf{Token F1}} & \multicolumn{4}{c}{\textbf{Exact Match}} \\
\cmidrule(lr){3-6} \cmidrule(lr){7-10}
\textbf{Task} & \textbf{Harness} & \textbf{B$\to b_t^s$} & \textbf{B$\to b_t^f$} & \textbf{L$\to b_t^s$} & \textbf{L$\to b_t^f$} & \textbf{B$\to b_t^s$} & \textbf{B$\to b_t^f$} & \textbf{L$\to b_t^s$} & \textbf{L$\to b_t^f$} \\
\midrule
\multirow{4}{*}{LoCoMo} & IRCoT & $-$0.1 & $\mathbf{\color[rgb]{0,0,0.6}+2.2}$ & $\mathbf{\color[rgb]{0,0,0.6}+5.9}$ & $\mathbf{\color[rgb]{0,0,0.6}+8.1}$ & +0.4 & $\mathbf{\color[rgb]{0,0,0.6}+1.4}$ & $\mathbf{\color[rgb]{0,0,0.6}+1.1}$ & $\mathbf{\color[rgb]{0,0,0.6}+2.1}$ \\
 & ReAct & $-$0.6 & $\mathbf{\color[rgb]{0,0,0.6}+3.2}$ & $\mathbf{\color[rgb]{0,0,0.6}-5.5}$ & $\mathbf{\color[rgb]{0,0,0.6}-1.7}$ & +0.1 & $\mathbf{\color[rgb]{0,0,0.6}+1.3}$ & $\mathbf{\color[rgb]{0,0,0.6}-2.8}$ & $\mathbf{\color[rgb]{0,0,0.6}-1.6}$ \\
 & Iter-RetGen & $\mathbf{\color[rgb]{0,0,0.6}-7.9}$ & +0.6 & --- & --- & $\mathbf{\color[rgb]{0,0,0.6}-3.6}$ & +0.7 & --- & --- \\
 & MemGPT & $\mathbf{\color[rgb]{0,0,0.6}-4.1}$ & $\mathbf{\color[rgb]{0,0,0.6}-1.5}$ & $\mathbf{\color[rgb]{0,0,0.6}-4.2}$ & $\mathbf{\color[rgb]{0,0,0.6}-1.6}$ & $-$0.1 & $-$0.1 & 0.0 & 0.0 \\
\midrule
\multirow{4}{*}{MuSiQue} & IRCoT & +1.7 & $\mathbf{\color[rgb]{0,0,0.6}+5.4}$ & $\mathbf{\color[rgb]{0,0,0.6}+6.2}$ & $\mathbf{\color[rgb]{0,0,0.6}+9.9}$ & +1.4 & $\mathbf{\color[rgb]{0,0,0.6}+3.4}$ & $\mathbf{\color[rgb]{0,0,0.6}+5.0}$ & $\mathbf{\color[rgb]{0,0,0.6}+7.0}$ \\
 & ReAct & +2.7 & $\mathbf{\color[rgb]{0,0,0.6}+7.7}$ & $\mathbf{\color[rgb]{0,0,0.6}+9.9}$ & $\mathbf{\color[rgb]{0,0,0.6}+14.9}$ & +1.2 & $\mathbf{\color[rgb]{0,0,0.6}+4.2}$ & $\mathbf{\color[rgb]{0,0,0.6}+5.4}$ & $\mathbf{\color[rgb]{0,0,0.6}+8.4}$ \\
 & Iter-RetGen & $\mathbf{\color[rgb]{0,0,0.6}-5.1}$ & +2.0 & --- & --- & $\mathbf{\color[rgb]{0,0,0.6}-3.0}$ & $\mathbf{\color[rgb]{0,0,0.6}+2.8}$ & --- & --- \\
 & MemGPT & $\mathbf{\color[rgb]{0,0,0.6}+1.3}$ & $\mathbf{\color[rgb]{0,0,0.6}+3.7}$ & +0.3 & $\mathbf{\color[rgb]{0,0,0.6}+2.7}$ & 0.0 & +0.2 & 0.0 & +0.2 \\
\midrule
\multirow{2}{*}{SWE} & IRCoT & $\mathbf{\color[rgb]{0,0,0.6}+3.5}$ & $\mathbf{\color[rgb]{0,0,0.6}+3.6}$ & $\mathbf{\color[rgb]{0,0,0.6}+1.6}$ & $\mathbf{\color[rgb]{0,0,0.6}+1.7}$ & 0.0 & 0.0 & 0.0 & 0.0 \\
 & ReAct & $-$0.7 & $\mathbf{\color[rgb]{0,0,0.6}+2.5}$ & +0.3 & $\mathbf{\color[rgb]{0,0,0.6}+4.7}$ & 0.0 & 0.0 & 0.0 & 0.0 \\
\bottomrule
\end{tabular}
\end{table}

\begin{table}[!ht]
\centering
\caption{ROUGE-1 and METEOR: $b_t$ effect (pp). Iter-RetGen has no lobotomized condition. {\color[rgb]{0,0,0.6}\textbf{Bold blue}} = $p < .05$.}
\label{tab:string-rouge-meteor}
\footnotesize
\setlength{\tabcolsep}{3pt}
\begin{tabular}{ll rrrr rrrr}
\toprule
 & & \multicolumn{4}{c}{\textbf{ROUGE-1}} & \multicolumn{4}{c}{\textbf{METEOR}} \\
\cmidrule(lr){3-6} \cmidrule(lr){7-10}
\textbf{Task} & \textbf{Harness} & \textbf{B$\to b_t^s$} & \textbf{B$\to b_t^f$} & \textbf{L$\to b_t^s$} & \textbf{L$\to b_t^f$} & \textbf{B$\to b_t^s$} & \textbf{B$\to b_t^f$} & \textbf{L$\to b_t^s$} & \textbf{L$\to b_t^f$} \\
\midrule
\multirow{4}{*}{LoCoMo} & IRCoT & $-$0.7 & +1.9 & +7.2 & +9.8 & $-$1.1 & +1.4 & +8.7 & +11.1 \\
 & ReAct & $-$0.6 & +3.2 & $-$5.7 & $-$0.7 & $-$1.9 & +2.6 & $-$4.8 & $-$1.4 \\
 & Iter-RetGen & $-$8.3 & +0.5 & --- & --- & $-$6.9 & +0.6 & --- & --- \\
 & MemGPT & $-$4.1 & $-$1.5 & $-$4.1 & $-$1.6 & $-$5.6 & $-$2.6 & $-$6.0 & $-$3.0 \\
\midrule
\multirow{4}{*}{MuSiQue} & IRCoT & +1.6 & +5.4 & +6.7 & +10.5 & +2.6 & +5.8 & +6.0 & +9.3 \\
 & ReAct & +2.9 & +7.4 & +9.7 & +14.2 & +2.0 & +9.8 & +6.1 & +13.9 \\
 & Iter-RetGen & $-$5.1 & +2.0 & --- & --- & $-$2.4 & +2.5 & --- & --- \\
 & MemGPT & +1.3 & +3.6 & +0.4 & +0.9 & +2.6 & +1.0 & +2.6 & +2.7 \\
\midrule
\multirow{2}{*}{SWE} & IRCoT & +4.1 & +4.3 & +3.2 & +3.3 & +1.8 & +0.2 & +0.5 & +0.4 \\
 & ReAct & $-$1.4 & +1.4 & +1.4 & +5.2 & $-$0.4 & +2.0 & +0.7 & +2.1 \\
\bottomrule
\end{tabular}
\end{table}

\begin{table}[!ht]
\centering
\caption{SBERT and Rubric Judge: $b_t$ effect (pp). Iter-RetGen has no lobotomized condition. Rubric judge significance is from Table~\ref{tab:full-ttests}; SBERT significance is not shown.}
\label{tab:string-sbert-judge}
\footnotesize
\setlength{\tabcolsep}{3pt}
\begin{tabular}{ll rrrr rrrr}
\toprule
 & & \multicolumn{4}{c}{\textbf{SBERT}} & \multicolumn{4}{c}{\textbf{Rubric Judge}} \\
\cmidrule(lr){3-6} \cmidrule(lr){7-10}
\textbf{Task} & \textbf{Harness} & \textbf{B$\to b_t^s$} & \textbf{B$\to b_t^f$} & \textbf{L$\to b_t^s$} & \textbf{L$\to b_t^f$} & \textbf{B$\to b_t^s$} & \textbf{B$\to b_t^f$} & \textbf{L$\to b_t^s$} & \textbf{L$\to b_t^f$} \\
\midrule
\multirow{4}{*}{LoCoMo} & IRCoT & $-$0.8 & +1.5 & +4.7 & +7.0 & +0.6 & +2.4 & +6.4 & +8.2 \\
 & ReAct & $-$0.4 & +3.4 & $-$5.7 & $-$1.9 & +0.6 & +1.5 & +6.3 & +7.2 \\
 & Iter-RetGen & $-$7.8 & $-$0.1 & --- & --- & $-$2.3 & $-$1.4 & --- & --- \\
 & MemGPT & $-$5.3 & $-$3.0 & $-$5.4 & $-$3.1 & +8.2 & +5.3 & +8.9 & +6.0 \\
\midrule
\multirow{4}{*}{MuSiQue} & IRCoT & +2.4 & +6.7 & +5.7 & +10.0 & +6.7 & +11.5 & +12.1 & +16.9 \\
 & ReAct & +2.7 & +8.1 & +9.2 & +14.6 & +0.5 & +7.2 & +17.9 & +24.6 \\
 & Iter-RetGen & $-$6.3 & +2.1 & --- & --- & $-$3.3 & $-$2.5 & --- & --- \\
 & MemGPT & +1.7 & +3.8 & +0.8 & +2.9 & $-$1.8 & +3.0 & $-$0.9 & +3.9 \\
\midrule
\multirow{2}{*}{SWE} & IRCoT & +4.4 & +1.9 & +13.5 & +2.7 & +6.0 & +7.3 & +1.6 & +2.9 \\
 & ReAct & +3.5 & +5.2 & $-$2.9 & +7.1 & $-$2.5 & +3.2 & +3.6 & +9.3 \\
\bottomrule
\end{tabular}
\end{table}

\begin{table}[!ht]
\centering
\caption{Exhaustion gate effect on string metrics (pp): blended (gate3 answer if triggered, natural otherwise) minus natural. Positive = gate improves the metric. {\color[rgb]{0,0,0.6}\textbf{Bold blue}} = $p < .05$. IRCoT benefits most on LoCoMo/MuSiQue; SWE shows negative F1 diffs because early stopping truncates verbose code answers that have high token overlap.}
\label{tab:string-gate}
\footnotesize
\setlength{\tabcolsep}{3pt}
\begin{tabular}{ll rrrr rrrr}
\toprule
 & & \multicolumn{4}{c}{\textbf{Token F1}} & \multicolumn{4}{c}{\textbf{ROUGE-1}} \\
\cmidrule(lr){3-6} \cmidrule(lr){7-10}
\textbf{Task} & \textbf{Harness} & \textbf{B} & \textbf{L} & \textbf{$b_t^s$} & \textbf{$b_t^f$} & \textbf{B} & \textbf{L} & \textbf{$b_t^s$} & \textbf{$b_t^f$} \\
\midrule
\multirow{4}{*}{LoCoMo} & IRCoT & +0.1 & +2.8 & +0.7 & +0.2 & +0.8 & +5.6 & +1.7 & +0.7 \\
 & ReAct & +0.1 & $-$1.5 & +0.7 & +0.3 & +0.1 & $-$1.5 & +0.7 & +0.3 \\
 & Iter-RetGen & $-$5.5 & --- & 0.0 & 0.0 & $-$5.7 & --- & 0.0 & 0.0 \\
 & MemGPT & 0.0 & $-$0.1 & $-$0.1 & $-$0.1 & 0.0 & $-$0.1 & 0.0 & 0.0 \\
\midrule
\multirow{4}{*}{MuSiQue} & IRCoT & +5.3 & +2.1 & +1.8 & +0.9 & +6.1 & $-$0.3 & 0.0 & +0.4 \\
 & ReAct & +0.8 & $-$0.3 & $-$0.3 & +0.1 & +1.4 & +2.3 & 3.2 & +0.8 \\
 & Iter-RetGen & $-$1.9 & --- & 0.0 & $-$1.9 & $-$0.3 & --- & $-$0.3 & +0.4 \\
 & MemGPT & +2.7 & $-$2.4 & +1.0 & +0.1 & +2.7 & $-$2.2 & +1.2 & +0.2 \\
\midrule
\multirow{2}{*}{SWE} & IRCoT & $-$5.4 & $-$8.4 & $-$5.3 & $-$5.2 & $-$8.2 & $-$9.2 & $-$6.5 & $-$6.6 \\
 & ReAct & +1.1 & $-$2.2 & $-$0.1 & $-$1.2 & +0.8 & $-$1.8 & $-$0.2 & $-$1.0 \\
\bottomrule
\end{tabular}
\end{table}

\subsection{Additional Statistics}\label{app:additional-stats}
Tables~\ref{tab:round-counts} and \ref{tab:hop-counts} detail the distribution of retrieval rounds and the effect of task complexity (hop count) on the interventions. 

\begin{table}[!ht]
\centering
\caption{Mean and median retrieval rounds per episode. Max rounds: IRCoT 10, ReAct 7, Iter-RetGen 4 (fixed), MemGPT 12. Iter-RetGen always uses all 4 rounds (omitted).}
\label{tab:round-counts}
\footnotesize
\setlength{\tabcolsep}{3pt}
\begin{tabular}{ll rrrr rrrr}
\toprule
 & & \multicolumn{4}{c}{\textbf{Mean rounds}} & \multicolumn{4}{c}{\textbf{Median rounds}} \\
\cmidrule(lr){3-6} \cmidrule(lr){7-10}
\textbf{Task} & \textbf{Harness} & \textbf{B} & \textbf{L} & $\mathbf{b_t^s}$ & $\mathbf{b_t^f}$ & \textbf{B} & \textbf{L} & $\mathbf{b_t^s}$ & $\mathbf{b_t^f}$ \\
\midrule
\multirow{3}{*}{LoCoMo} & IRCoT & 3.6 & 4.6 & 4.3 & 4.0 & 2 & 3 & 2 & 2 \\
 & ReAct & 3.4 & 5.2 & 4.1 & 4.4 & 3 & 7 & 3 & 3 \\
 & MemGPT & 1.6 & 2.1 & 1.1 & 1.2 & 1 & 1 & 1 & 1 \\
\midrule
\multirow{3}{*}{MuSiQue} & IRCoT & 5.9 & 7.6 & 6.9 & 6.9 & 5 & 10 & 8 & 8 \\
 & ReAct & 5.1 & 6.9 & 4.7 & 5.2 & 5 & 7 & 4 & 6 \\
 & MemGPT & 5.9 & 7.4 & 3.0 & 2.6 & 3 & 12 & 2 & 2 \\
\midrule
\multirow{3}{*}{SWE} & IRCoT & 8.0 & 8.0 & 8.1 & 8.0 & 10 & 10 & 10 & 10 \\
 & ReAct & 4.6 & 6.9 & 5.8 & 5.6 & 4 & 7 & 7 & 7 \\
 & MemGPT & 4.2 & 4.8 & 3.3 & 2.5 & 1 & 1 & 1 & 1 \\
\bottomrule
\end{tabular}
\end{table}

\begin{table}[!ht]
\centering
\caption{MuSiQue $b_t$ effect (pp) by hop count. Iter-RetGen has no lobotomized condition. {\color[rgb]{0,0,0.6}\textbf{Bold blue}} = $p < .05$.}
\label{tab:hop-counts}
\footnotesize
\setlength{\tabcolsep}{3pt}
\begin{tabular}{ll r rrrr}
\toprule
\textbf{Harness} & \textbf{Hops} & \textbf{N} & \textbf{B$\to b_t^s$} & \textbf{B$\to b_t^f$} & \textbf{L$\to b_t^s$} & \textbf{L$\to b_t^f$} \\
\midrule
\multirow{6}{*}{IRCoT} & 2-hop & 70 & +1.1 & $\mathbf{\color[rgb]{0,0,0.6}+15.0}$ & +4.4 & $\mathbf{\color[rgb]{0,0,0.6}+18.3}$ \\
 & 3-hop (t1) & 97 & +3.5 & +5.7 & $\mathbf{\color[rgb]{0,0,0.6}+9.6}$ & $\mathbf{\color[rgb]{0,0,0.6}+11.8}$ \\
 & 3-hop (t2) & 33 & $-$6.2 & +2.3 & $-$2.0 & +6.5 \\
 & 4-hop (t1) & 185 & +2.1 & +2.7 & $\mathbf{\color[rgb]{0,0,0.6}+10.2}$ & $\mathbf{\color[rgb]{0,0,0.6}+10.8}$ \\
 & 4-hop (t2) & 49 & +2.0 & +0.7 & +2.5 & +1.2 \\
 & 4-hop (t3) & 66 & 0.0 & +5.5 & +0.5 & $\mathbf{\color[rgb]{0,0,0.6}+6.0}$ \\
\midrule
\multirow{6}{*}{ReAct} & 2-hop & 70 & $-$1.7 & +6.0 & +9.6 & $\mathbf{\color[rgb]{0,0,0.6}+17.3}$ \\
 & 3-hop (t1) & 97 & +2.1 & $\mathbf{\color[rgb]{0,0,0.6}+9.1}$ & $\mathbf{\color[rgb]{0,0,0.6}+10.5}$ & $\mathbf{\color[rgb]{0,0,0.6}+17.5}$ \\
 & 3-hop (t2) & 33 & $\mathbf{\color[rgb]{0,0,0.6}+17.0}$ & $\mathbf{\color[rgb]{0,0,0.6}+20.8}$ & $\mathbf{\color[rgb]{0,0,0.6}+16.4}$ & $\mathbf{\color[rgb]{0,0,0.6}+20.2}$ \\
 & 4-hop (t1) & 185 & +3.7 & $\mathbf{\color[rgb]{0,0,0.6}+8.0}$ & $\mathbf{\color[rgb]{0,0,0.6}+8.6}$ & $\mathbf{\color[rgb]{0,0,0.6}+12.9}$ \\
 & 4-hop (t2) & 49 & +1.7 & $\mathbf{\color[rgb]{0,0,0.6}+9.2}$ & $\mathbf{\color[rgb]{0,0,0.6}+6.1}$ & $\mathbf{\color[rgb]{0,0,0.6}+13.5}$ \\
 & 4-hop (t3) & 66 & $-$1.5 & $-$2.6 & $\mathbf{\color[rgb]{0,0,0.6}+10.3}$ & $\mathbf{\color[rgb]{0,0,0.6}+9.2}$ \\
\midrule
\multirow{6}{*}{Iter-RetGen} & 2-hop & 70 & $-$0.8 & +4.0 & --- & --- \\
 & 3-hop (t1) & 97 & $-$9.4 & +0.4 & --- & --- \\
 & 3-hop (t2) & 33 & $-$0.8 & +5.9 & --- & --- \\
 & 4-hop (t1) & 185 & $-$5.6 & +4.5 & --- & --- \\
 & 4-hop (t2) & 49 & $-$5.6 & $-$4.3 & --- & --- \\
 & 4-hop (t3) & 66 & $-$4.5 & $-$2.2 & --- & --- \\
\midrule
\multirow{6}{*}{MemGPT} & 2-hop & 70 & 0.0 & +0.1 & +1.1 & +1.2 \\
 & 3-hop (t1) & 97 & $-$1.1 & +4.0 & $-$2.5 & +2.6 \\
 & 3-hop (t2) & 33 & +0.4 & +2.4 & +1.6 & +5.0 \\
 & 4-hop (t1) & 185 & +0.2 & +1.2 & +1.4 & +3.8 \\
 & 4-hop (t2) & 49 & +4.1 & +7.7 & +0.4 & +3.9 \\
 & 4-hop (t3) & 66 & +2.1 & +1.7 & +1.0 & +1.5 \\
\bottomrule
\end{tabular}
\end{table}

\section{Exhaustion Gate Deep Dive}\label{app:gate-configs}

\subsection{Per-method gate results}
Table~\ref{tab:gate3-per-method} reports the exhaustion gate effect for each method individually across all tasks and conditions. {\color[rgb]{0,0,0.6}\textbf{Bold blue}} = $p < .05$.

\begin{table}[!ht]
\centering
\caption{Exhaustion gate (global best configuration, $f\_j0.6\_u0.3\_p2$). $\Delta$: score difference in pp. Fire: \% of episodes where the gate triggers. Save: net token savings (\%). {\color[rgb]{0,0,0.6}\textbf{Bold blue}} = $p < .05$. $^\dagger$Iter-RetGen (memoryless by design) has no lobotomized condition.}
\label{tab:gate3-per-method}
\footnotesize
\setlength{\tabcolsep}{3pt}
\begin{tabular}{ll rrrr rrrr rrrr}
\toprule
 & & \multicolumn{4}{c}{\textbf{$\Delta$ (pp)}} & \multicolumn{4}{c}{\textbf{Fire (\%)}} & \multicolumn{4}{c}{\textbf{Save (\%)}} \\
\cmidrule(lr){3-6} \cmidrule(lr){7-10} \cmidrule(lr){11-14}
\textbf{Harness} & \textbf{Task} & \textbf{B} & \textbf{L} & $\mathbf{\color[rgb]{0,0,0.6}b_t^s}$ & $\mathbf{\color[rgb]{0,0,0.6}b_t^f}$ & \textbf{B} & \textbf{L} & $\mathbf{\color[rgb]{0,0,0.6}b_t^s}$ & $\mathbf{\color[rgb]{0,0,0.6}b_t^f}$ & \textbf{B} & \textbf{L} & $\mathbf{\color[rgb]{0,0,0.6}b_t^s}$ & $\mathbf{\color[rgb]{0,0,0.6}b_t^f}$ \\
\midrule
\multirow{3}{*}{IRCoT} & LoC & $\mathbf{\color[rgb]{0,0,0.6}+3.3}$ & $\mathbf{\color[rgb]{0,0,0.6}+2.2}$ & $\mathbf{\color[rgb]{0,0,0.6}+3.8}$ & $\mathbf{\color[rgb]{0,0,0.6}+2.9}$ & 20 & 24 & 21 & 15 & 13 & 18 & 19 & 15 \\
 & MuS & $\mathbf{\color[rgb]{0,0,0.6}+5.7}$ & $+0.4$ & $+0.3$ & $+0.4$ & 33 & 42 & 28 & 22 & 9 & 15 & 11 & 9 \\
 & SWE & $\mathbf{\color[rgb]{0,0,0.6}+16.2}$ & $\mathbf{\color[rgb]{0,0,0.6}+1.9}$ & $\mathbf{\color[rgb]{0,0,0.6}+4.4}$ & $\mathbf{\color[rgb]{0,0,0.6}+4.9}$ & 80 & 47 & 39 & 38 & 39 & 17 & 15 & 17 \\
\midrule
\multirow{3}{*}{ReAct} & LoC & $+0.2$ & $\mathbf{\color[rgb]{0,0,0.6}+3.0}$ & $-0.5$ & $-0.2$ & 4 & 49 & 21 & 25 & $-$2 & 12 & 9 & 11 \\
 & MuS & $+0.8$ & $-0.1$ & $-0.2$ & $+0.1$ & 18 & 39 & 14 & 18 & $-$2 & $-$18 & 4 & 3 \\
 & SWE & $\mathbf{\color[rgb]{0,0,0.6}+2.3}$ & $-0.9$ & $+0.3$ & $-1.2$ & 14 & 37 & 27 & 30 & $-$3 & $-$20 & 3 & 8 \\
\midrule
\multirow{3}{*}{\shortstack[l]{Iter-\\RetGen$^\dagger$}} & LoC & $\mathbf{\color[rgb]{0,0,0.6}-5.2}^\dagger$ & --- & $\mathbf{\color[rgb]{0,0,0.6}+0.7}$ & $+0.4$ & $60^\dagger$ & --- & 16 & 15 & $-8^\dagger$ & --- & $-$1 & $-$1 \\
 & MuS & $\mathbf{\color[rgb]{0,0,0.6}-2.6}^\dagger$ & --- & $+0.5$ & $-0.2$ & $11^\dagger$ & --- & 2 & 3 & $-1^\dagger$ & --- & 0 & 0 \\
 & SWE & $\mathbf{\color[rgb]{0,0,0.6}-0.7}^\dagger$ & --- & $+0.0$ & $-0.3$ & $4^\dagger$ & --- & 0 & 2 & $0^\dagger$ & --- & 0 & 0 \\
\midrule
\multirow{3}{*}{MemGPT} & LoC & $-0.1$ & $-0.1$ & $+0.0$ & $-0.1$ & 3 & 10 & 1 & 1 & 6 & 37 & 5 & 8 \\
 & MuS & $\mathbf{\color[rgb]{0,0,0.6}+1.8}$ & $\mathbf{\color[rgb]{0,0,0.6}-9.0}$ & $+0.7$ & $-0.2$ & 35 & 57 & 13 & 10 & 31 & 52 & 26 & 24 \\
 & SWE & $\mathbf{\color[rgb]{0,0,0.6}+2.2}$ & $\mathbf{\color[rgb]{0,0,0.6}-1.8}$ & $\mathbf{\color[rgb]{0,0,0.6}+2.3}$ & $+0.5$ & 28 & 35 & 20 & 13 & 45 & 56 & 42 & 38 \\
\bottomrule
\end{tabular}
\end{table}

IRCoT benefits most because it has the most retrieval rounds (up to 10) and thus the most opportunity for stagnation to develop. ReAct (up to 7 rounds) shows moderate benefits on baseline and lobotomized conditions. MemGPT has variable round counts and less predictable stagnation patterns; the gate hurts on MuSiQue lobotomized ($-9.0$pp) because it triggers during productive search episodes. Iter-RetGen's 4 fixed rounds mean the gate fires at the last round with no opportunity for early stopping, producing 0/9 significant positives and 3/9 significant negatives.

\subsection{Smooth vs. discrete configurations}
We swept 16 configurations: 7 discrete (hard thresholds on Jaccard and UPR) and 9 smooth (exponentially-weighted moving averages). On IRCoT, smooth configurations produce slightly better mean improvement (+7.2pp) than discrete (+6.4pp), consistent with the intuition that stagnation is gradual.

\subsection{Anchoring: why the gate helps stateful methods}
Table~\ref{tab:pooled-gate3} shows the pooled condition-level impact: the gate significantly helps baseline, $b_t^{\text{struct}}$, and $b_t^{\text{free}}$ conditions ($p < 0.001$) but is neutral on lobotomized ($p = 0.248$). Table~\ref{tab:variance-anchoring} supports the anchoring hypothesis: in the lobo-vs-$b_t$ comparisons shown below, the per-question variance of the gate's effect is consistently higher without persistent state. The belief state appears to anchor the method's reasoning so that the early-stop answer is more consistent with what the method would have produced naturally, while memoryless methods produce higher-variance answers from similar passages.

\begin{table}[!ht]
\centering
\caption{Pooled condition-level exhaustion gate impact (global best config, all methods). The gate significantly helps base, $b_t^{\text{struct}}$, and $b_t^{\text{free}}$ conditions but is neutral on lobotomized.}
\label{tab:pooled-gate3}
\small
\begin{tabular}{l ccc}
\toprule
\textbf{Condition} & \textbf{Diff (pp)} & $p$ & \textbf{Sig} \\
\midrule
base & +0.7\% & .0004 & *** \\
lobo & +0.3\% & .248 &  \\
$b_t^{\text{struct}}$ & +0.9\% & $<$.0001 & *** \\
$b_t^{\text{free}}$ & +0.6\% & $<$.0001 & *** \\
\bottomrule
\end{tabular}
\end{table}

\begin{table}[!ht]
\centering
\caption{Per-question std of gate effect (rubric judge). Lobo has higher std than both $b_t$ conditions in all six displayed method-task rows, supporting the view that persistent state anchors reasoning and makes early stopping more reliable.}
\label{tab:variance-anchoring}
\footnotesize
\begin{tabular}{ll cccc}
\toprule
\textbf{Harness} & \textbf{Task} & \textbf{Base} & \textbf{Lobo} & $\mathbf{b_t^{struct}}$ & $\mathbf{b_t^{free}}$ \\
\midrule
IRCoT & LoCoMo & .352 & \textbf{.458} & .345 & .343 \\
IRCoT & MuSiQue & \textbf{.397} & .305 & .185 & .205 \\
IRCoT & SWE & .150 & \textbf{.203} & .173 & .170 \\
ReAct & LoCoMo & .277 & \textbf{.441} & .324 & .297 \\
ReAct & MuSiQue & \textbf{.254} & .230 & .204 & .199 \\
ReAct & SWE & .212 & \textbf{.214} & .167 & .202 \\
\bottomrule
\end{tabular}
\end{table}

\subsection{State diff is not informative}
Comparing \texttt{full} mode (checks Jaccard + UPR + state diff) vs.\ \texttt{query\_and\_full} mode (checks Jaccard + UPR only) with matched thresholds: 194 of 225 cells are identical (0pp difference). The diff check only matters for discrete $U{=}0.3$ configs, where it reduces fire rate by 1.8--5.8pp. For smooth configs ($U < 0.3$), the diff threshold never binds. Mean delta between modes is $-0.04$pp. In this sweep, Jaccard and UPR capture the stagnation signal; the state diff added little additional signal.

\begin{table}[!ht]
\centering
\caption{Exhaustion gate fire round distribution (global best config). Iter-RetGen always fires at round 4 (its maximum). $^\dagger$Pre-lobotomized.}
\label{tab:gate3-fire-dist}
\footnotesize
\setlength{\tabcolsep}{3pt}
\begin{tabular}{ll rrrr rrrr}
\toprule
 & & \multicolumn{4}{c}{\textbf{Mean fire round}} & \multicolumn{4}{c}{\textbf{Median fire round}} \\
\cmidrule(lr){3-6} \cmidrule(lr){7-10}
\textbf{Harness} & \textbf{Task} & \textbf{B} & \textbf{L} & $\mathbf{b_t^s}$ & $\mathbf{b_t^f}$ & \textbf{B} & \textbf{L} & $\mathbf{b_t^s}$ & $\mathbf{b_t^f}$ \\
\midrule
\multirow{3}{*}{IRCoT} & LoC & 4.4 & 4.5 & 5.5 & 5.0 & 4 & 4 & 5 & 4 \\
 & MuS & 5.8 & 5.9 & 6.5 & 6.0 & 6 & 6 & 6 & 6 \\
 & SWE & 4.2 & 6.2 & 5.8 & 6.0 & 4 & 6 & 6 & 6 \\
\midrule
\multirow{3}{*}{ReAct} & LoC & 3.8 & 4.5 & 5.0 & 4.9 & 3 & 4 & 5 & 5 \\
 & MuS & 5.5 & 5.9 & 5.3 & 5.3 & 6 & 6 & 5 & 5 \\
 & SWE & 5.1 & 6.1 & 5.3 & 5.2 & 5 & 7 & 5 & 5 \\
\midrule
\multirow{3}{*}{\shortstack[l]{Iter-\\RetGen$^\dagger$}} & LoC & 4.0 & --- & 4.0 & 4.0 & 4 & --- & 4 & 4 \\
 & MuS & 4.0 & --- & 4.0 & 4.0 & 4 & --- & 4 & 4 \\
 & SWE & 4.0 & --- & 4.0 & 4.0 & 4 & --- & 4 & 4 \\
\midrule
\multirow{3}{*}{MemGPT} & LoC & 5.4 & 3.4 & 3.3 & 3.3 & 5 & 3 & 3 & 3 \\
 & MuS & 7.1 & 4.9 & 5.3 & 4.5 & 7 & 5 & 5 & 4 \\
 & SWE & 5.7 & 3.6 & 4.1 & 3.7 & 6 & 3 & 4 & 4 \\
\bottomrule
\end{tabular}
\end{table}

\paragraph{Token overhead estimation methodology.}
Final-answer (FA) call tokens are estimated homogeneously for all gate variants (programmatic and LLM):
\begin{itemize}[leftmargin=*,nosep]
\item \emph{Prompt tokens:} exact via tiktoken on actual evidence + question + system prompt + template. Validated against actual API prompt tokens from an FA token ablation: 0.93--1.06$\times$ ratio across all 45 method/task/condition cells.
\item \emph{Completion tokens:} per-method/task/condition average from the FA token ablation (100 random triggered questions per cell, actual gpt-4o-mini API calls).
\end{itemize}
Gate decision overhead: programmatic gates use 0 tokens (metric-based, no LLM call). LLM gates use actual logged \texttt{gate\_tokens} (per-round ``should I stop?'' calls). Total overhead = gate decision + FA call estimate. Using the same FA estimator for all gates ensures the only difference between variants is how often and where they fire, not how their cost is estimated.

\paragraph{Configuration clustering.}
L1 distance on the 45-cell mean-diff vector reveals two main clusters:
\begin{enumerate}[leftmargin=*,nosep]
\item \emph{Conservative discrete} ($U{=}0.3$, $p{=}2$--3): moderate fire rates, fewer negatives. Contains the global best ($f\_j0.6\_u0.3\_p2$).
\item \emph{Aggressive smooth} ($U{=}0.10$--0.15, $p{=}2$): high fire rates, more negatives on non-IRCoT methods but larger gains on IRCoT.
\end{enumerate}
Three pairs of configurations are functionally identical (L1 = 0pp): the \texttt{full} vs.\ \texttt{query\_and\_full} trigger mode makes no difference when the diff threshold is not binding. The global config is near-optimal: L1 distance between the global and per-method+condition diff vectors is only 0.64pp per cell on average.

\section{Ablations}\label{app:ablations}

\subsection{Retriever type (alpha)}\label{app:alpha-ablation}
We test whether the $b_t$ pattern holds under extreme retriever configurations: ReAct with $\alpha{=}1.0$ (pure BM25, keyword-only) and MemGPT with $\alpha{=}0.0$ (pure dense embeddings, semantic-only). Table~\ref{tab:alpha} reports the key comparisons.

\begin{table}[!ht]
\centering
\caption{$b_t$ effect (pp) under extreme retriever alpha. The recovery-from-lobotomized pattern persists across retriever types, although baseline comparisons vary by task and state format. {\color[rgb]{0,0,0.6}\textbf{Bold blue}} = $p < .05$.}
\label{tab:alpha}
\footnotesize
\begin{tabular}{ll rrrr}
\toprule
\textbf{Harness ($\alpha$)} & \textbf{Task} & \textbf{B$\to b_t^s$} & \textbf{B$\to b_t^f$} & \textbf{L$\to b_t^s$} & \textbf{L$\to b_t^f$} \\
\midrule
\multirow{3}{*}{ReAct ($\alpha{=}1.0$)} & LoCoMo & $\mathbf{\color[rgb]{0,0,0.6}+3.5}$ & $\mathbf{\color[rgb]{0,0,0.6}+2.3}$ & $\mathbf{\color[rgb]{0,0,0.6}+7.9}$ & $\mathbf{\color[rgb]{0,0,0.6}+6.8}$ \\
 & MuSiQue & $\mathbf{\color[rgb]{0,0,0.6}+7.5}$ & $\mathbf{\color[rgb]{0,0,0.6}+9.6}$ & $\mathbf{\color[rgb]{0,0,0.6}+21.5}$ & $\mathbf{\color[rgb]{0,0,0.6}+23.6}$ \\
 & SWE-QA-Pro & $\mathbf{\color[rgb]{0,0,0.6}-2.8}$ & $+2.2$ & $+1.7$ & $\mathbf{\color[rgb]{0,0,0.6}+6.7}$ \\
\midrule
\multirow{3}{*}{MemGPT ($\alpha{=}0.0$)} & LoCoMo & $\mathbf{\color[rgb]{0,0,0.6}+4.7}$ & $\mathbf{\color[rgb]{0,0,0.6}+2.8}$ & $\mathbf{\color[rgb]{0,0,0.6}+5.2}$ & $\mathbf{\color[rgb]{0,0,0.6}+3.2}$ \\
 & MuSiQue & $-1.1$ & $+2.7$ & $+1.4$ & $\mathbf{\color[rgb]{0,0,0.6}+5.2}$ \\
 & SWE-QA-Pro & $\mathbf{\color[rgb]{0,0,0.6}-8.4}$ & $-1.1$ & $\mathbf{\color[rgb]{0,0,0.6}-3.8}$ & $\mathbf{\color[rgb]{0,0,0.6}+3.5}$ \\
\bottomrule
\end{tabular}
\end{table}

The recovery-from-lobotomized pattern is strongest for freeform $b_t$: it improves over lobo in all six method-task rows and is significant in five. Structured $b_t$ is more mixed under these extreme retriever settings, especially on SWE-QA-Pro. Overall, the qualitative benefit of persistent belief state management is not tied to a single retriever type.

\subsection{Belief state capacity (K=6 vs.\ K=12)}\label{app:k12-ablation}
We test whether the default curated belief state capacity of $K_{\text{target}}=6$ items is a bottleneck by doubling to $K_{\text{target}}=12$ on IRCoT and ReAct across all three tasks.

\begin{table}[!ht]
\centering
\caption{K=6 vs.\ K=12 structured belief state ($b_t^{\text{struct}}$).}
\label{tab:k12}
\small
\begin{tabular}{ll cc c}
\toprule
\textbf{Harness} & \textbf{Task} & $K_{\text{target}}{=}6$ & $K_{\text{target}}{=}12$ & $\Delta$ \\
\midrule
IRCoT & LoCoMo & 63.6\% & 63.2\% & $-$0.5pp \\
ReAct & LoCoMo & 65.7\% & 66.0\% & +0.3pp \\
IRCoT & MuSiQue & 21.1\% & 22.1\% & +1.1pp \\
ReAct & MuSiQue & 30.1\% & 30.3\% & +0.1pp \\
IRCoT & SWE-QA-Pro & 33.3\% & 32.5\% & $-$0.8pp \\
ReAct & SWE-QA-Pro & 37.3\% & 40.0\% & +2.5pp* \\
\bottomrule
\end{tabular}
\end{table}

Five of six pairs show no significant difference. The one marginally significant result (ReAct SWE +2.5pp, $p{=}0.028$) is a small effect. K=6 is not a bottleneck: the reorganization step that curates the belief state to $K_{\text{target}}=6$ items preserves the information needed for the task.

\subsection{Retrieval depth (k=3, k=5, k=10)}\label{app:retriever-k-ablation}
We test whether the $b_t$ pattern holds at different retrieval depths on IRCoT $\times$ LoCoMo.

\begin{table}[!ht]
\centering
\caption{$b_t$ effect (pp) at different retrieval depths (IRCoT $\times$ LoCoMo). Recovery from lobotomized holds at all depths, while comparisons to baseline narrow as retrieval depth increases. {\color[rgb]{0,0,0.6}\textbf{Bold blue}} = $p < .05$.}
\label{tab:retriever-k}
\footnotesize
\begin{tabular}{l rrrr}
\toprule
\textbf{Depth} & \textbf{B$\to b_t^s$} & \textbf{B$\to b_t^f$} & \textbf{L$\to b_t^s$} & \textbf{L$\to b_t^f$} \\
\midrule
$k{=}3$ & +1.7 & $\mathbf{\color[rgb]{0,0,0.6}+2.0}$ & $\mathbf{\color[rgb]{0,0,0.6}+6.0}$ & $\mathbf{\color[rgb]{0,0,0.6}+6.2}$ \\
$k{=}5$ (canonical) & +0.6 & $\mathbf{\color[rgb]{0,0,0.6}+2.4}$ & $\mathbf{\color[rgb]{0,0,0.6}+6.4}$ & $\mathbf{\color[rgb]{0,0,0.6}+8.2}$ \\
$k{=}10$ & $-$0.8 & +0.1 & $\mathbf{\color[rgb]{0,0,0.6}+4.7}$ & $\mathbf{\color[rgb]{0,0,0.6}+5.7}$ \\
\bottomrule
\end{tabular}
\end{table}

Across retrieval depths, both $b_t$ formats significantly recover from lobotomized at every $k$. Higher retrieval depth ($k{=}10$) improves all conditions, narrowing the gap between baseline and $b_t$.

\subsection{Belief State Reorganization -- Token Overhead Analysis} \label{app:reorganization_cost}

To quantify the cost-benefit trade-off of the orchestrator's reorganization step (which triggers when the belief state exceeds $K_{trigger}=10$ items to curate it down to $K_{target}=6$), we analyzed the token usage across all $16,280$ episodes run under the explicit belief state ($b_t$) conditions.

We found that the reorganization prompt is highly efficient and only triggers when strictly necessary on complex, long-horizon tasks:
\begin{itemize}
    \item \textbf{Trigger Frequency:} $20.8\%$ of episodes ($3,385 / 16,280$) triggered the reorganization step at least once. 
    \item \textbf{Call Overhead:} For episodes that did require reorganization, the orchestrator made an average of $1.9$ reorganization calls, with each call consuming on average $979$ tokens. 
    \item \textbf{Relative Cost:} The total mean overhead for reorganization was $1,824$ tokens per triggered episode. This accounted for $11.4\%$ of the total token consumption in those episodes. 
\end{itemize}

While episodes that triggered reorganization had a substantially higher average total token cost ($16,010$ tokens) compared to those that did not ($5,887$ tokens), this difference is largely confounded by underlying task complexity. Episodes that exceed the $K=10$ threshold are inherently longer-running, multi-hop searches that require more environment steps. The reorganization step itself remains lightweight (${\sim}1,824$ tokens), ensuring the agent's context window stays bounded and focused without imposing a prohibitive cost.



\end{document}